\documentclass[letter,conference]{IEEEtran}
\usepackage{fancyhdr}
\usepackage{graphicx,times,amsmath} 
\usepackage{subfigure,comment,enumerate,morefloats,float, siunitx}
\usepackage{amssymb}
\usepackage{nicefrac}
\usepackage{xspace,multirow}

\IEEEoverridecommandlockouts
  {\begin{tabbing}\hspace*{0.5cm}\=\hspace*{0.5cm}\=\hspace*{0.5cm}\=\hspace*{0.5cm}\=\hspace*{0.5cm}\=\hspace*{0.5cm}\=\hspace*{0.5cm}\=\hspace*{0.5cm}\kill}%
  {\end{tabbing}}

\begin{document}
\title{An Investigation into the Relationship Between Type-2 FOU Size and Environmental Uncertainty in Robotic Control}
\author{\IEEEauthorblockN{Naisan Benatar, Uwe Aickelin and Jonathan M. Garibaldi {\it Member, IEEE}}
\IEEEauthorblockA{Intelligent Modelling and Analysis Research Group \\School of Computer Science\\
University of Nottingham\\
Email: [nxb,uxa,jmg]@cs.nott.ac.uk}
}
\maketitle

\author{Naisan Benatar, Uwe Aickelin\vspace{-4ex} and Jonathan M. Garibaldi {\it Member, IEEE} } 

\maketitle

\thispagestyle{fancy}
\fancyhead{}
\lhead{}
\lfoot{}
\cfoot{}
\rfoot{}
\renewcommand{\headrulewidth}{0pt}
\renewcommand{\footrulewidth}{0pt}

\begin{abstract}

It has been suggested that, when faced with large amounts of uncertainty in situations of automated control, type-2 fuzzy logic based controllers will out-perform the simpler type-1 varieties due to the latter lacking the flexibility to adapt accordingly.  This paper aims to investigate this problem in detail in order to analyse when a type-2 controller will improve upon type-1 performance.  A robotic sailing boat is subjected to several experiments in which the uncertainty and  difficulty of the sailing problem is increased in order to observe the effects on measured performance.  Improved performance is observed but not in every case.  The size of the FOU is shown to be have a large effect on performance with potentially severe performance penalties for incorrectly sized footprints.

\textit{Keywords: Interval Type-2 Fuzzy, Robot Boat control, Fuzzy Control, Uncertainty}
\end{abstract}

\section{Introduction}

A fuzzy logic system maps inputs into a fuzzy set by means of a \textit{fuzzifier}.  The fuzzy set output is then processed as part of a Fuzzy Inference System (FIS) where, the set is used as an input to an inference system with its associated rule base.  This results in a new output set that can itself be \textit{defuzzified} into a value suitable for use as a standard (e.g. PID) controller output.

The way the processing of the fuzzy sets is handled can be varied based upon application specific requirements or restrictions and this gives rise to three main subcategories of fuzzy control: type-1, interval type-2 and general type-2.  In this paper we consider type-1 and interval type-2 based systems and apply both to control problems of increasing difficulty with increasing quantities of uncertainty.  We thereby hope to develop a method of determining parameters, such as FOU size that will give performance improvements over type-1 based controllers.  

	The application discussed in this paper is one of robotic sailing using the FLOATs (Fuzzy Logic Operated AuTonomous Sailboat) platform as described in \cite{Benatar2011}, in which a robotic sail boat actuates sail and rudder positions based on received sensor data with the aim of steering an autonomous boat around a predetermined course.  Similar boat based applications have been investigated with a variety of approaches such as PID \cite{Sauze2005}, biologically inspired \cite{Sauze2010} and fuzzy methods as in \cite{Stelzer-20081}. This application was selected due to the multiple sources of noise and uncertainty in the environment.  This maybe useful for discerning under which conditions type-2 based controllers will outperform type-1.

This paper is organised as follows:  Section \ref{sec:background} provides background and information into the methods and systems used in the rest of the paper.  Section \ref{sec:SEM} describes the experiments that will be performed and will be followed by Section \ref{sec:Results} where we state our numerical results and Section \ref{sec:Discussion} where these results are discussed.  Finally in Section \ref{sec:Conclusions_Future} we draw some conclusions along with some ideas for future work.

\section{Background}
\label{sec:background}

\subsection{Type-1 Fuzzy Logic}
\label{sec:t1}
   
\begin{figure*}[t]
\subfigure[Type-1 Membership function]{\label{fig:T1MF}\includegraphics[scale=0.3]{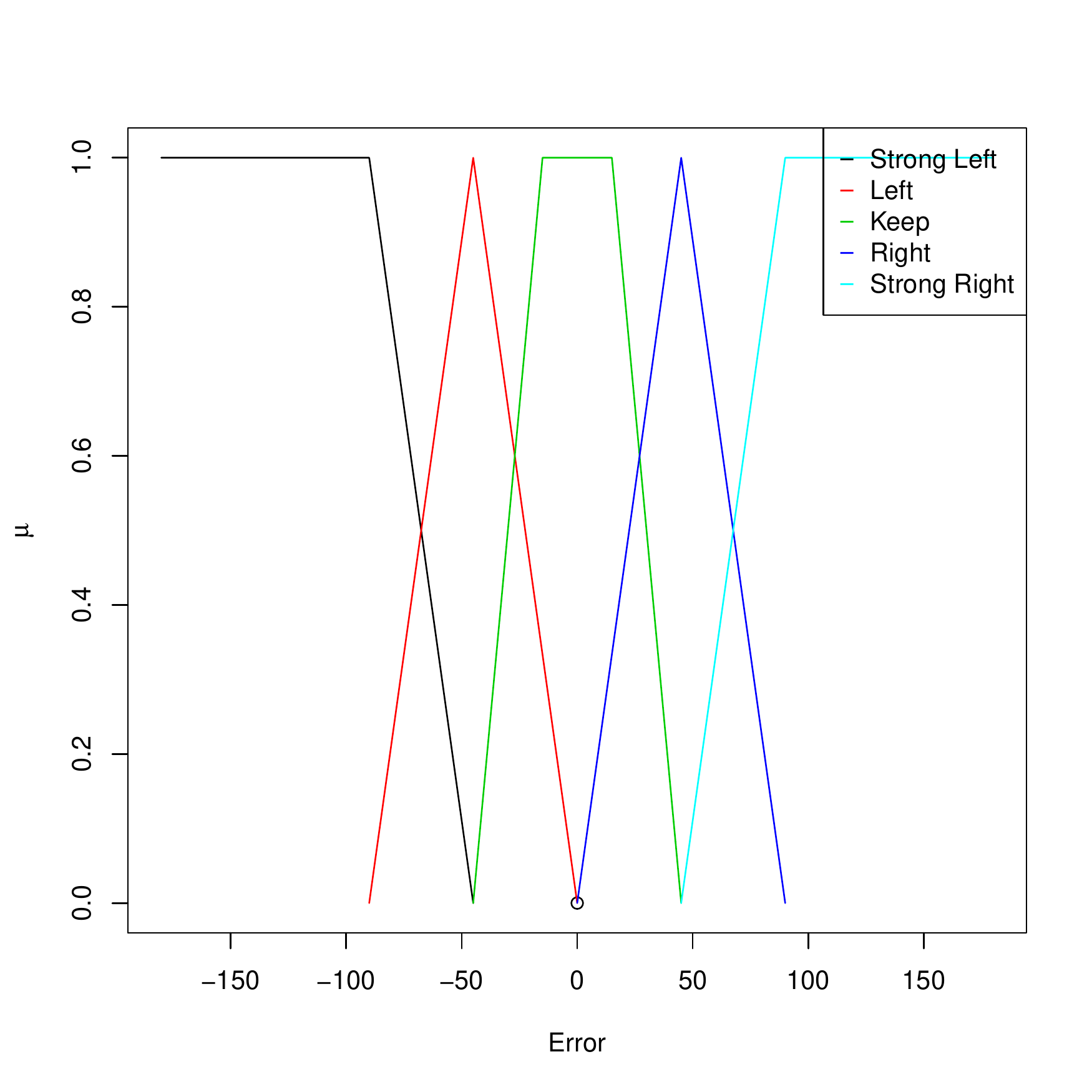}}
\subfigure[Movement 10 Type-2]{\label{fig:Movement10T2}\includegraphics[scale=0.3]{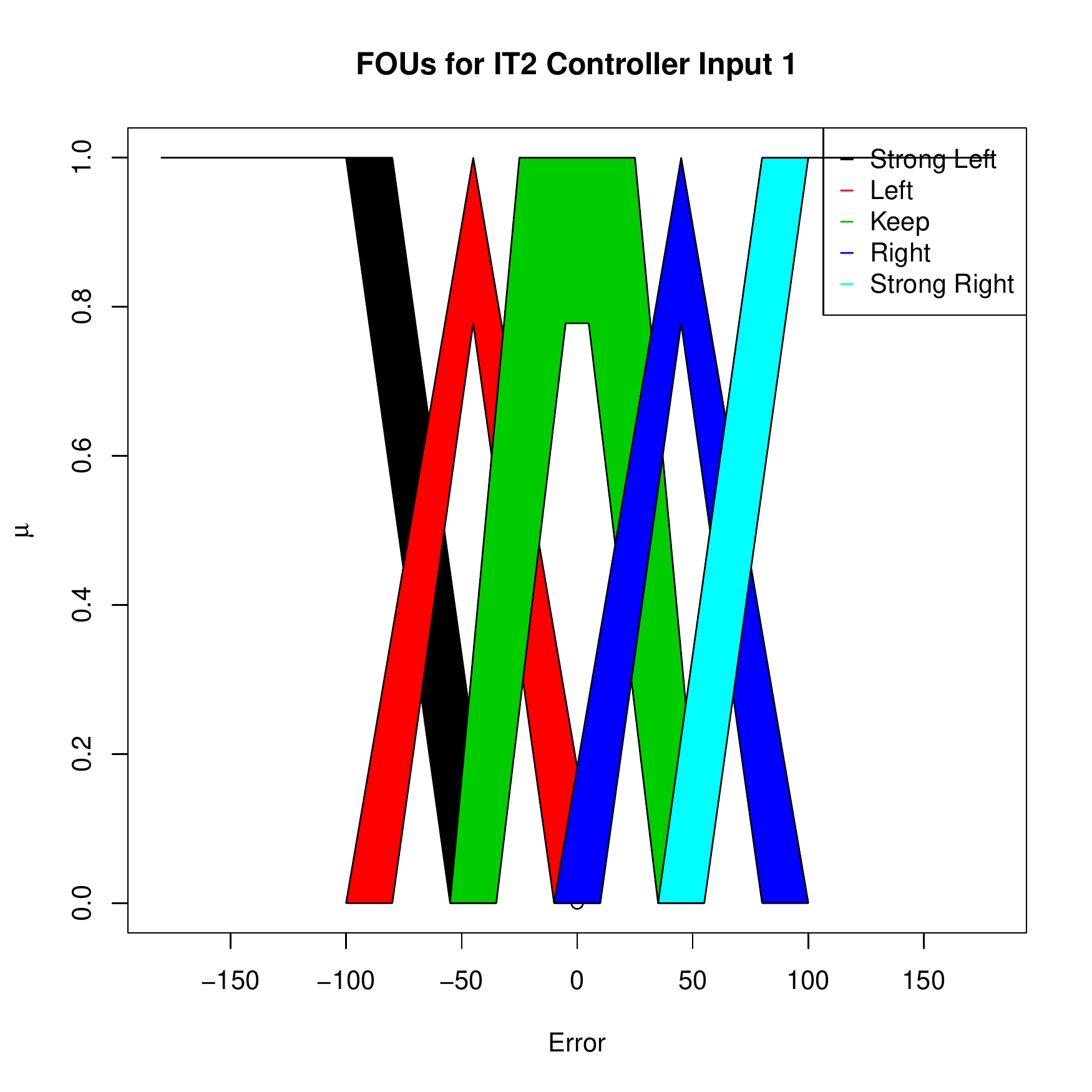}}
\subfigure[Movement 20 Type-2]{\label{fig:Movement20T2}\includegraphics[scale=0.3]{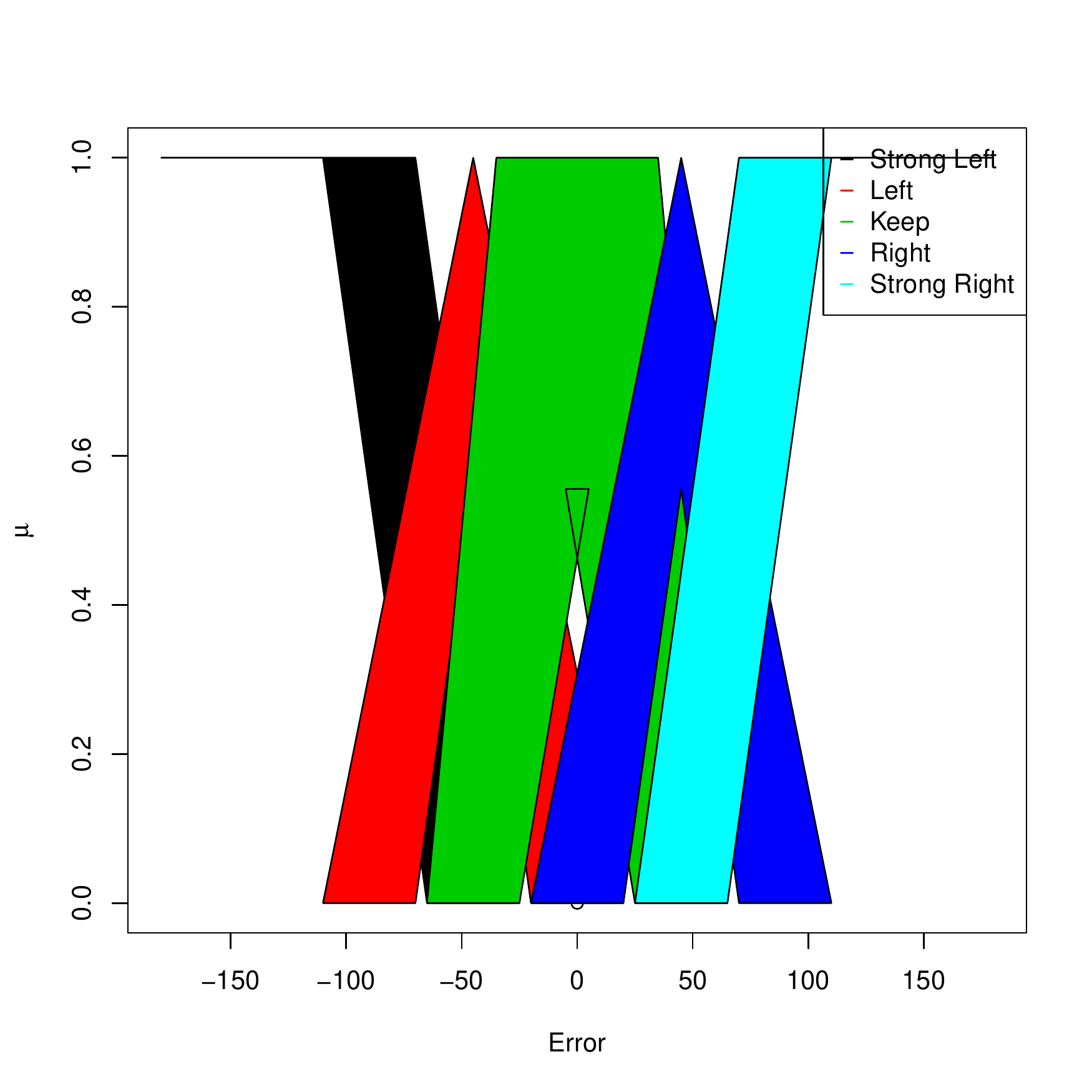}}
\caption{Membership functions of Fuzzy Controllers. }
\label{fig:Move10T2MF}
\end{figure*}
      
	Type-1 fuzzy logic, introduced by Zadeh in \cite{Zadeh1975} uses 2D membership functions that are commonly triangular or Gaussian in shape.  The $x$ axis represents the set of possible inputs into the system  as obtained from the application. The $y$ axis represents the degree to which the given input is a member of the fuzzy set and may have a value between 0 (no membership to the set) and 1 (complete membership to the set).  
	
	One of the main shortcomings of type-1 fuzzy logic is its limited flexibility. Moving a type-1 controller tuned in a specific environment into a different environment can often lead to significant performance degradation unless controller changes are made.  It has been stated by Mendel \cite{NileshNKarnik1999} that type-2 control may be advantageous over type-1 in many areas such as robust control in which the information to be processed is uncertain.  
	
\subsection{Type-2 Fuzzy Logic}
\label{sec:t2}

	  Type-2 fuzzy logic is a development of type-1 and with this development comes greater freedom and flexibility as discussed by Wu \cite{Wu2011a}. This may increase performance in certain situations as explored by Hagras \cite{hagras2004}, at the potential cost of increased computational load.  Some of the increased flexibility of this type of system can be attributed to the fact that the membership functions in type-2 systems are represented in three dimensions instead of type-1's two.  This extra dimension gives rise to what is known as the secondary membership function.  
	  
	  In an interval type-2 system the value of the secondary function is binary, allowing interval type-2 systems to be represented by two individual membership functions in the 2D plane.  They are termed the upper and lower membership functions (UMF and LMF respectivly).  The area that they enclose is termed the footprint of uncertainty (FOU).
	  
	  With general type-2 fuzzy logic, unlike in interval systems, the secondary membership function is continuous instead of discrete, for a more comprehensive overview of type-2 fuzzy logic the reader is directed at a paper such as \cite{Mendel2006}.
	  
	As a type-2 inference system will return at least 1 type-2 fuzzy set and this cannot be directly defuzzified a process known as \textit{type-reduction} is required . This process reduces this type-2 set into type-1 which can then be defuzzified into a usable output.
   
   While both interval and general type-2 are more complex than type-1, methods to reduce processing requirements are being developed which bring the possibility of type-2 based mobile robotic systems closer.  For example Wu \cite{Wu2006} and Wagner and Hagras \cite{Wagner2011} both present distinct methods to reduce the computational load significantly so that is is manageable by resource constrained systems such as those used on mobile robots.

\section{Experimental Methodology}
\label{sec:SEM}

\subsection{Problem description}

	The application to be used as our test case is FLOATS, described in \cite{Benatar2011}.  It is an autonomous sailing boat originally based on the work done by Stelzer in \cite{Stelzer-2008} which can work in both simulation and the real world environments. Wind, location, waypoint location and direction sensors are used to calculate error and delta error values.  These are used as inputs to a fuzzy inference system which produces a rudder change output value.
	
	In this paper we have opted to use simulation as the method of experimentation, with real life experimentation planned for future work.  The simulator used, Tracksail has been used by others for development and testing of autonomous sailing robot systems including Sauze and Neal \cite{Sauze200811}.  Tracksail is Java based open source software and communicates with boat controllers by means of a standard network socket.
	
	A running rate of 1Hz was fixed in the controller code for all controller configurations.  This value was chosen in order to ensure that the more sophisticated controllers could run a complete cycle as there were initial concerns that for type-2 based controllers especially the overhead would be too high for a faster rate.  While this low running rate will lead to overall lower performance we believe consistency between controllers is more important in this work.  A comparison of type-1 and type-2 controllers running at different speeds is reserved for future work.
	
	The cumulative RMSE (Root-mean square error) between the current and desired heading (measured in degrees) will be the main metric used for comparison. As is usual with control experiments, lower values represent a better performing controller.	 
	
\subsection{Hypothesis}
\label{sec:hyp}

	We hypothesize that there will be A point at which the difficulty of the course is sufficient that the type-2 will significantly improve upon the results of the standard type-1 controller.  We expect that this behaviour will occur more obviously in situations with a higher uncertainty score as described in Table \ref{tab:Configs} and with the extra turns exaggerating the effect further.   The uncertainty score is used only for giving an arbitrary ordering for the configurations and is calculated by summing the direction and speed uncertainty scores together. 
	
    It is expected that as the various wind configurations are tested the RMSE will increase in a predictable manner --- with configurations A and B showing lower RMSE values than the configurations H and I for example.  We do not expect a linear increase especially as several configurations have the same uncertainty score.
    
	We also expect to see that as the FOU size is increased the performance will start at type-1 levels. We expect this to be followed by an increase in performance and a subsequent drop as the FOU increases to cover more of the universe of discourse. This will result in extremely bad performance which, in the worst cases will prevent the course from being completed at all.	 

\subsection{Experimental Design}
\label{sec:Methodology}

	The controllers under test maintained the same membership functions throughout all experiments, with the type-1 membership functions shown in \ref{fig:T1MF}.  The only change in each run was the horizontal movement of the type-2 controller that alters the size of the FOU.  We tested six different values for each FOU size, starting at 0 and increasing to a maximum of 25 in increments of 5. An example FOU size 10 is shown in Figure \ref{fig:Movement10T2}.
	
	In our application we derive our type-2 footprints of uncertainty by introducing a horizontal movement to the type-1 membership functions, with the amount of movement being used will be varied as a parameter value.  Figures \ref{fig:Movement10T2} and \ref{fig:Movement20T2} show examples of this with the type-1 membership function being moved 10 and 20 respectively to give the shown FOUs.

	Two separate mechanisms were used to gradually increase the difficulty of completing the course, with the aim of highlighting the differences in performance that can be achieved with the various fuzzy systems under test.  The first mechanism is the way point system of the simulator which will allow us to define the number and size of turn that the controller must steer the boat through in order to complete the course.  The second mechanism is the introduction of noise into the environment in the form of variations of the wind.  Table \ref{tab:Windspeed} outlines the configurations of wind that we will be using in this experiment.
	
\begin{table}[tb]
\caption{Wind Speed and Direction Upper and Lower Values and Uncertainty Score}
\begin{tabular}{lrrr}
\hline\hline
Direction & \multicolumn{1}{l}{Uncertainty Score} & \multicolumn{1}{l}{Lower Limit} & \multicolumn{1}{l}{Upper Limit} \\ \hline
None & 0 & 180 & 180 \\ 
Low & 1 & 160 & 200 \\ 
High & 2 & 140 & 220 \\ 
&&& \\ 
\hline\hline
Speed & \multicolumn{1}{l}{Uncertainty Score} & \multicolumn{1}{l}{Lower Limit} & \multicolumn{1}{l}{Upper Limit} \\ 
None & 0 & 7 & 7 \\ 
Low & 1 & 4 & 10 \\ 
High & 2 & 1 & 13 \\ 
\hline\hline
\label{tab:Windspeed}
\end{tabular}
\caption{Wind Configurations of Experiments.  The Uncertainty score is the sum of the directional and speed uncertainty scores as shown in Table \ref{tab:Windspeed}.}
\begin{tabular}{lllr}
\hline\hline
Configuration & Speed  & Direction  & \multicolumn{1}{l}{Uncertainty } \\
& Uncertainty & Uncertainty & Score \\\hline
A & None & None & 0 \\ 
B & Low & None & 1 \\ 
C & None & Low & 1 \\ 
D & Low & Low & 2 \\ 
E & High & None & 2 \\ 
F & None & High & 2 \\ 
G & High  & Low & 3 \\ 
H & Low & High & 3 \\ 
I & High & High & 4 \\ 
\hline\hline
\end{tabular}
\label{tab:Configs}

\end{table}
	
	An automated control rig was used to execute batches of 30 runs for each combination of controller, parameter value (FOU size) and course layout. Each piece of software (controller, simulator and common code) maintains its own logging files that can be analysed to produce RMSE values and other interesting statistics.

\subsection{Course Design}

\begin {figure}[t]
		\includegraphics [scale=0.3]{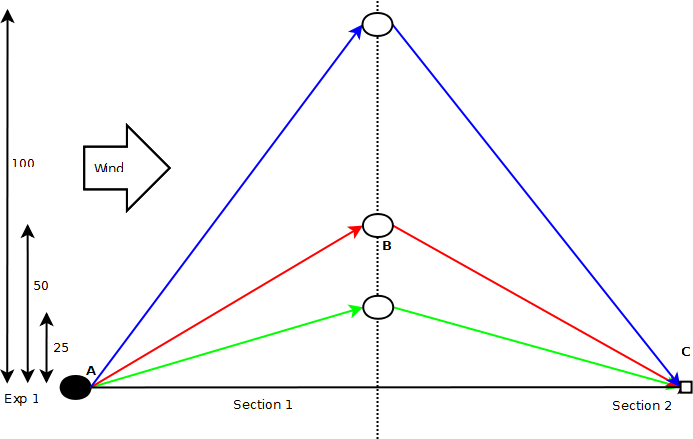}
		\caption{Each coloured line represents a course layout under test.  The white circles represent end points and the black circle the start point. The angles required for the first turn are 5.71$^{\circ}$, 11.42$^{\circ}$ and 21.84$^{\circ}$ for 25, 50 and 100 meters vertical movements respectively.  Not to scale.}
 		\label{fig:Course}

\end {figure}

	The courses will be built up from the simplest of all courses --- a straight line with a parallel fixed wind, in which the boat must simply move forward in order to complete the course.  The difficulty will then be increased by adding turns of varying angles as shown in Figure \ref{fig:Course}. It can be observed that the courses under test will containing either one or two turns.  The vertical movement required to complete the course, defined as either 0, 25, 50 or 100 metres, will alter the angle that the boat must turn in order to complete the course.  The angle required for the first turn are 5.71$^{\circ}$, 11.42$^{\circ}$ and 21.84$^{\circ}$ for 25, 50 and 100 meters vertical movements respectively while the second turn, being twice as large will be 11.4$^{\circ}$, 22.84$^{\circ}$ and 43.68$^{\circ}$.

	Every combination of course layout and wind configurations, as shown in Table \ref{tab:Configs}, will be tested with each controller configuration.  We will start with no noise (configuration A) and move towards the most uncertain environment (Configuration I). Every four seconds a wind change will be triggered by the simulator using a Gaussian random number generator to change the values of the wind speed and direction.  This gradual increase of noise will allow structured observations to be made about the effects of noise upon the performance of type-2 controllers with varying FOU sizes.
	
\section{Results}
\label{sec:Results} 

One sided Wilcoxon tests were used to test the statistical significance of the difference between two individual batches of experiments.   The result for this test is a P-Value with a small value ($<$0.0005) indicating a statistically significant difference.  For clarity course layouts are displayed as the vertical distance hyphenated with the number of turns, for example Single-25 would indicate a course which a single turn and 25m of vertical movement.

The first test for all experiments was a comparison of the type-1 controller metric values with the FOU size 0 type-2 controller values in order to ensure the values of each were statistically similar.  This allowed us to ensure all parts of the simulator setup were functioning correctly and gave a good sanity check for each experiment.

\begin{figure}[bt]
\centering
\label{fig:0_Point}
\subfigure[Single-0 RMSE Change]{\label{fig:0_Config}\includegraphics[scale=0.3] {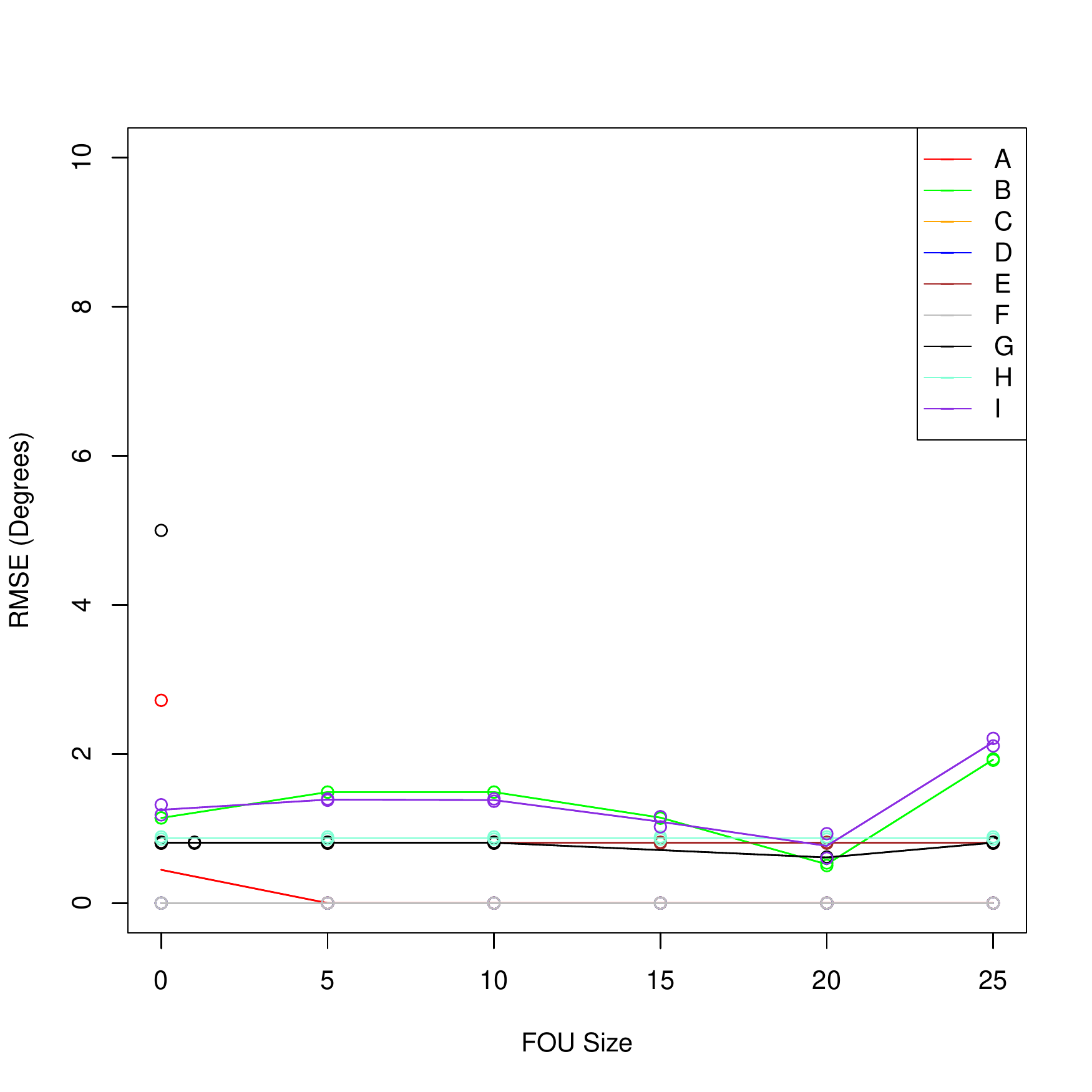}}\\
\subfigure[Single-0 RMSE Difference]{\label{fig:0_Gain}\includegraphics[scale = 0.3]{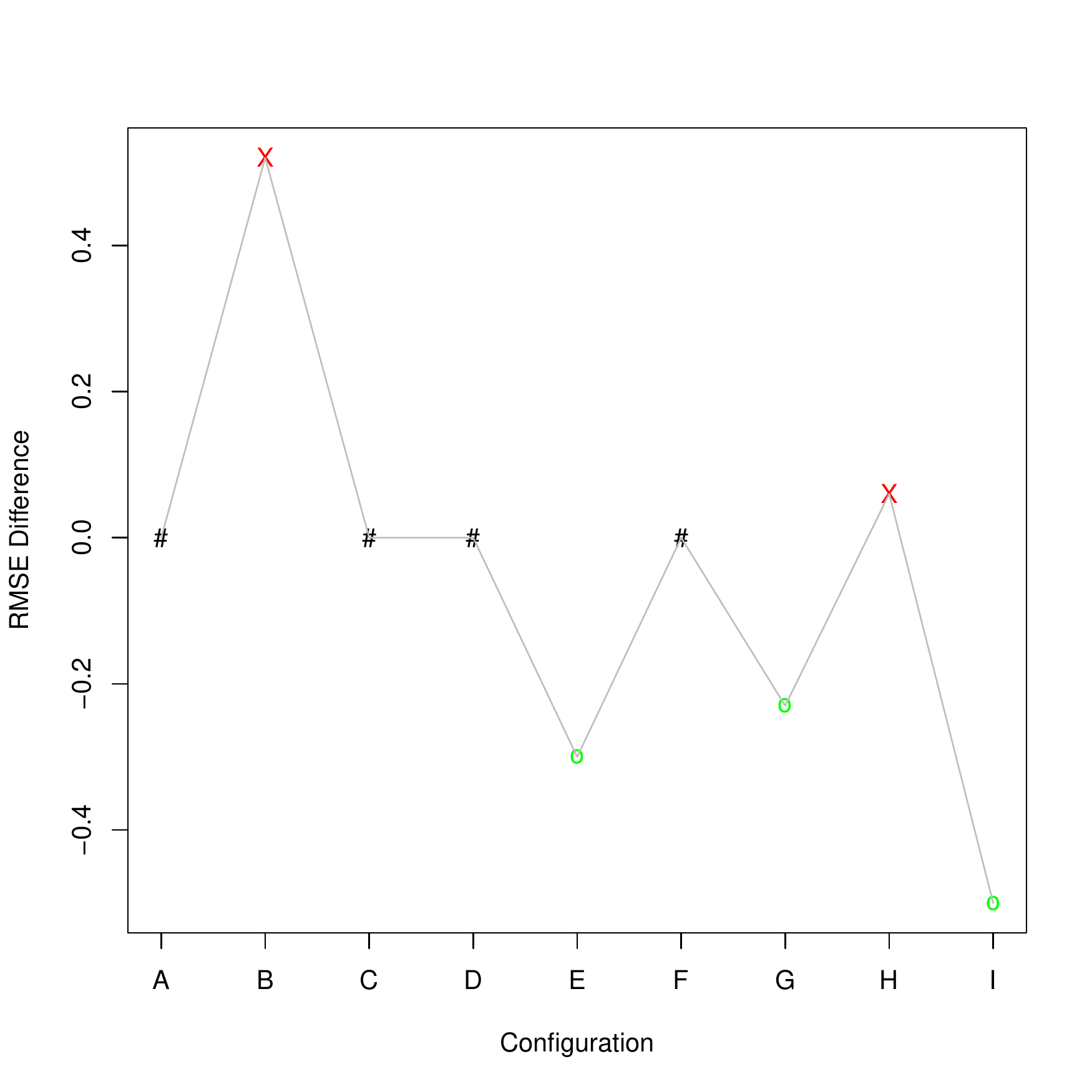}}\\
\subfigure[Single-0 Course Plot]{\label{fig:0_Plot}\includegraphics[scale = 0.3]{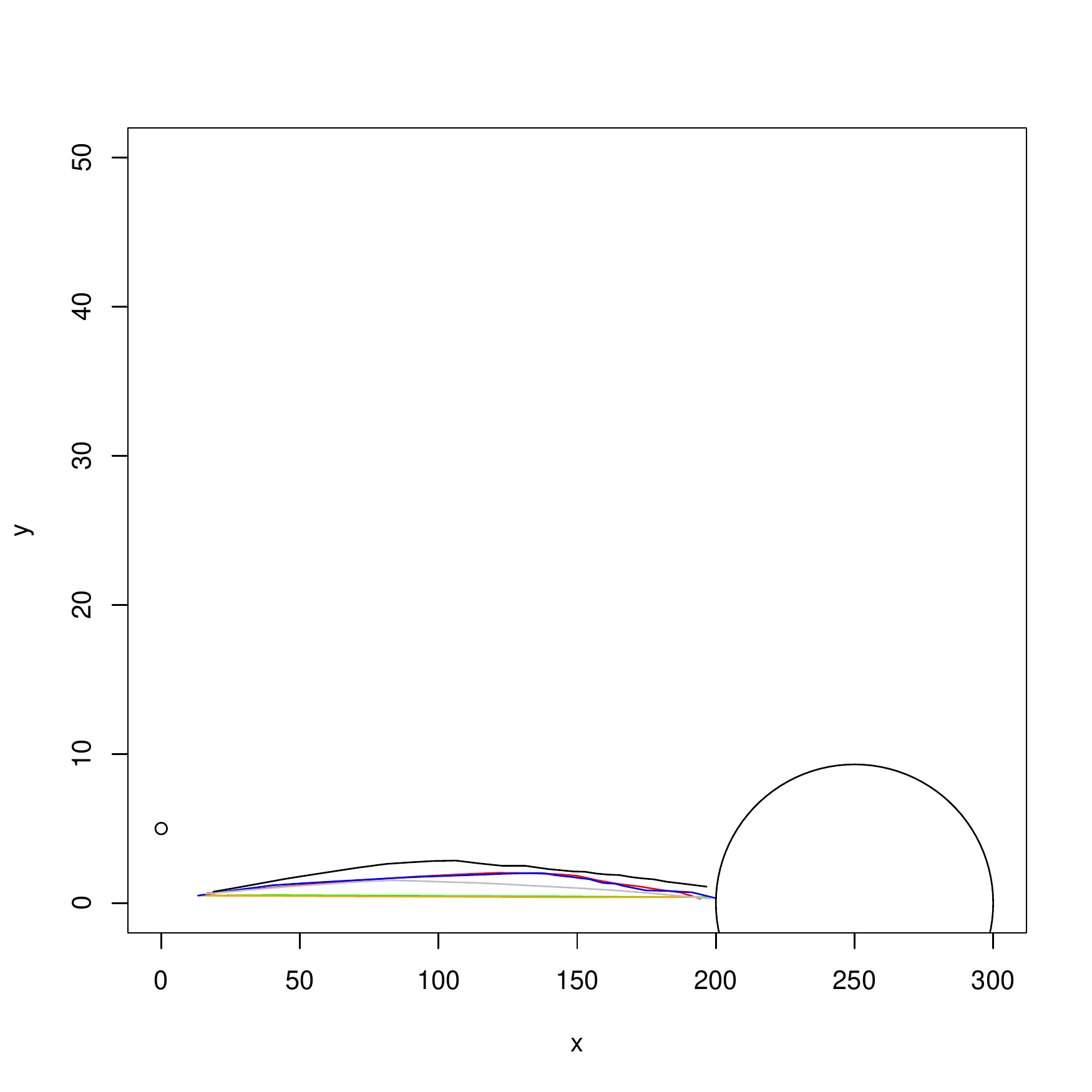}}

\caption{Benchmark Experiment Results}
\end{figure}

\begin{table}[tb]
\caption{RMSE Difference between Type-1 and a Type-2 Controller with FOU size of 20 on Single-50 course layout.  This increase in performance can also be observed in Figure \ref{fig:50_Config}}
\begin{tabular}{lrrr}
\hline\hline
Configuration & \multicolumn{1}{l}{Type-1 RMSE} & \multicolumn{1}{l}{Type-2 RMSE} & \multicolumn{1}{l}{RMSE Difference} \\ 
\hline
A & 5.93 & 3.56 & -2.37 \\ 
B & 8.35 & 3.91 & -4.44 \\ 
C & 6.34 & 3.31 & -3.03 \\ 
D & 5.90 & 3.20 & -2.70 \\ 
E & 7.41 & 4.08 & -3.33 \\ 
F & 4.83 & 2.84 & -1.99 \\ 
G & 6.32 & 3.46 & -2.86 \\ 
H & 5.10 & 2.44 & -2.66 \\ 
I & 4.72 & 2.66 & -2.06 \\ 
\hline\hline
\label{tab:50_2_Diff}
\end{tabular}

\caption{RMSE Difference between Type-1 and a Type-2 Controller with FOU size of 20 on Single-100 course layout}
\begin{tabular}{lrrr}
\hline\hline
Configuration & \multicolumn{1}{l}{Type-1 RMSE} & \multicolumn{1}{l}{Type-2 RMSE} & \multicolumn{1}{l}{RMSE Difference} \\ \hline
A & 15.29 & 15.70 & 0.41 \\ 
B & 15.75 & 22.43 & 6.69 \\ 
C & 11.84 & 16.68 & 4.83 \\ 
D & 12.33 & 17.15 & 4.82 \\ 
E & 12.53 & 25.68 & 13.15 \\ 
F & 11.67 & 15.53 & 3.86 \\ 
G & 14.53 & 15.50 & 0.97 \\ 
H & 13.68 & 22.22 & 8.54 \\ 
I & 12.97 & 16.35 & 3.38 \\ 
\hline\hline
\end{tabular}
\label{tab:100_2_Point}
\end{table}

\begin{figure*}[t]
\centering
\subfigure[25m Vertical Movement]{\label{fig:25_Config}\includegraphics[scale=0.3] {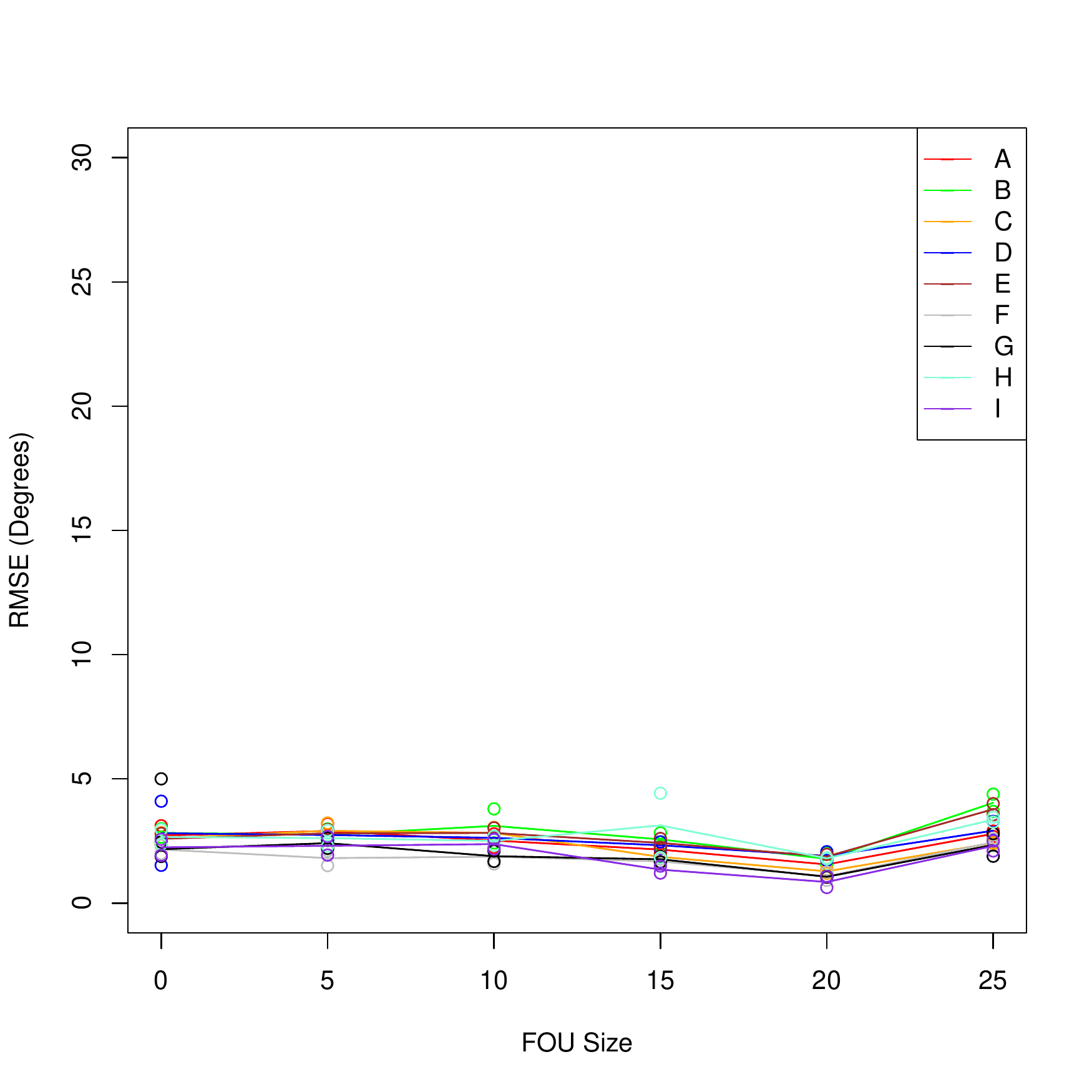}}
\subfigure[50m Vertical Movement]{\label{fig:50_Config}\includegraphics[scale=0.3]{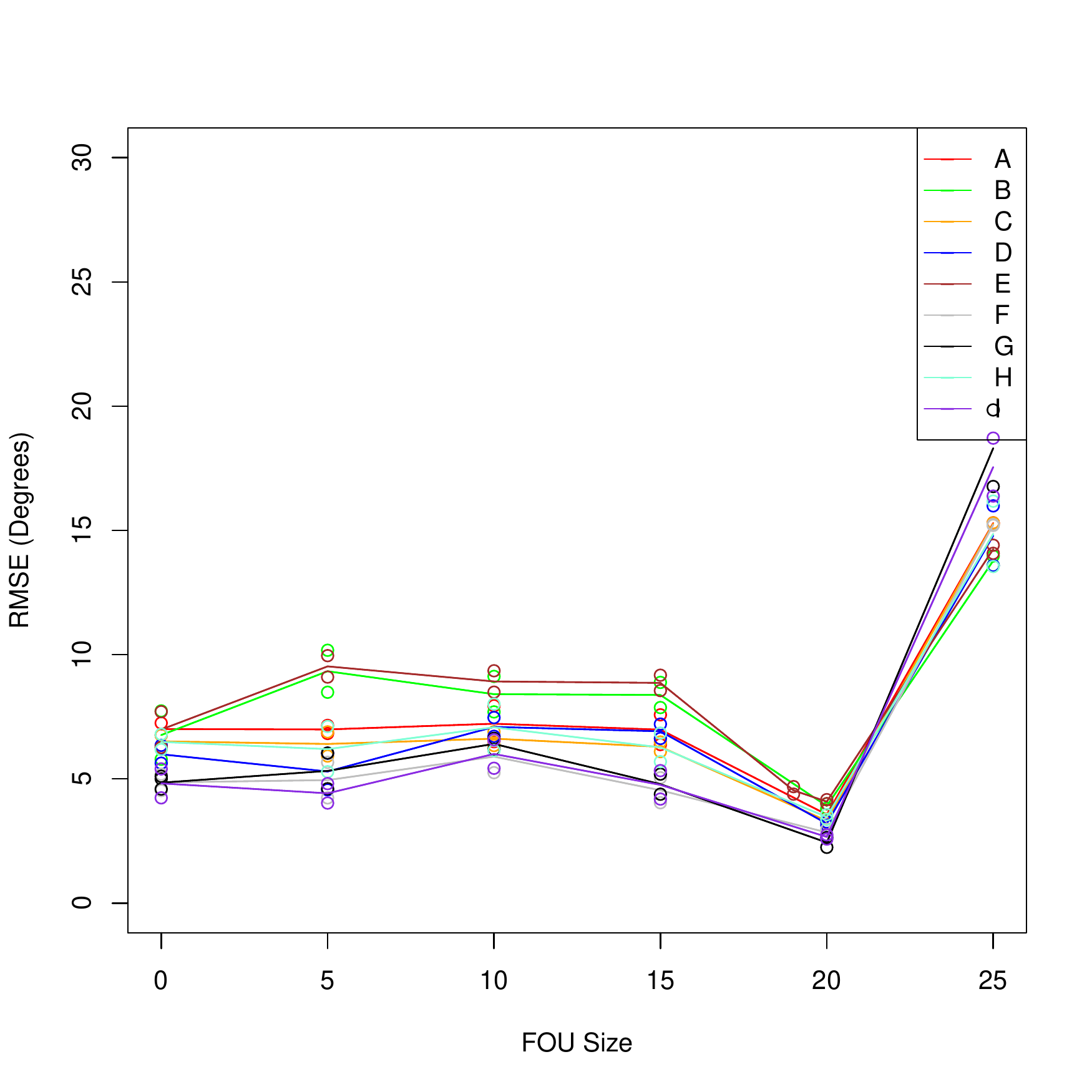}}
\subfigure[100m Vertical Movement]{\label{fig:100_Config}\includegraphics[scale=0.3]{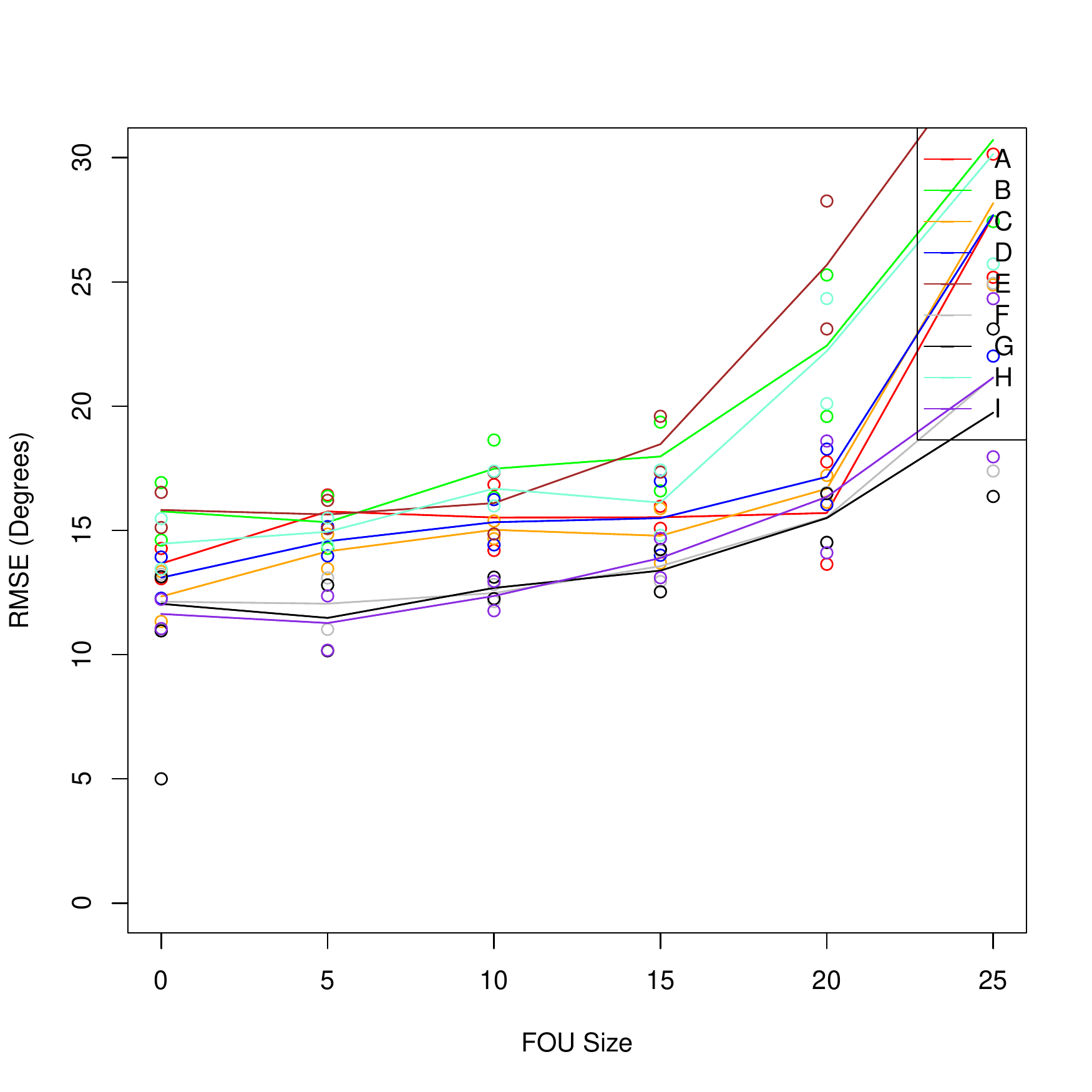}}
\caption{Single Turn Experiments RMSE Changes as FOU size increased}
\label{fig:2_Config}
\end{figure*}

\begin{figure*}[t]
\centering
\subfigure[25m Vertical Movement]{\label{fig:3_0_Config}\includegraphics[scale=0.3] {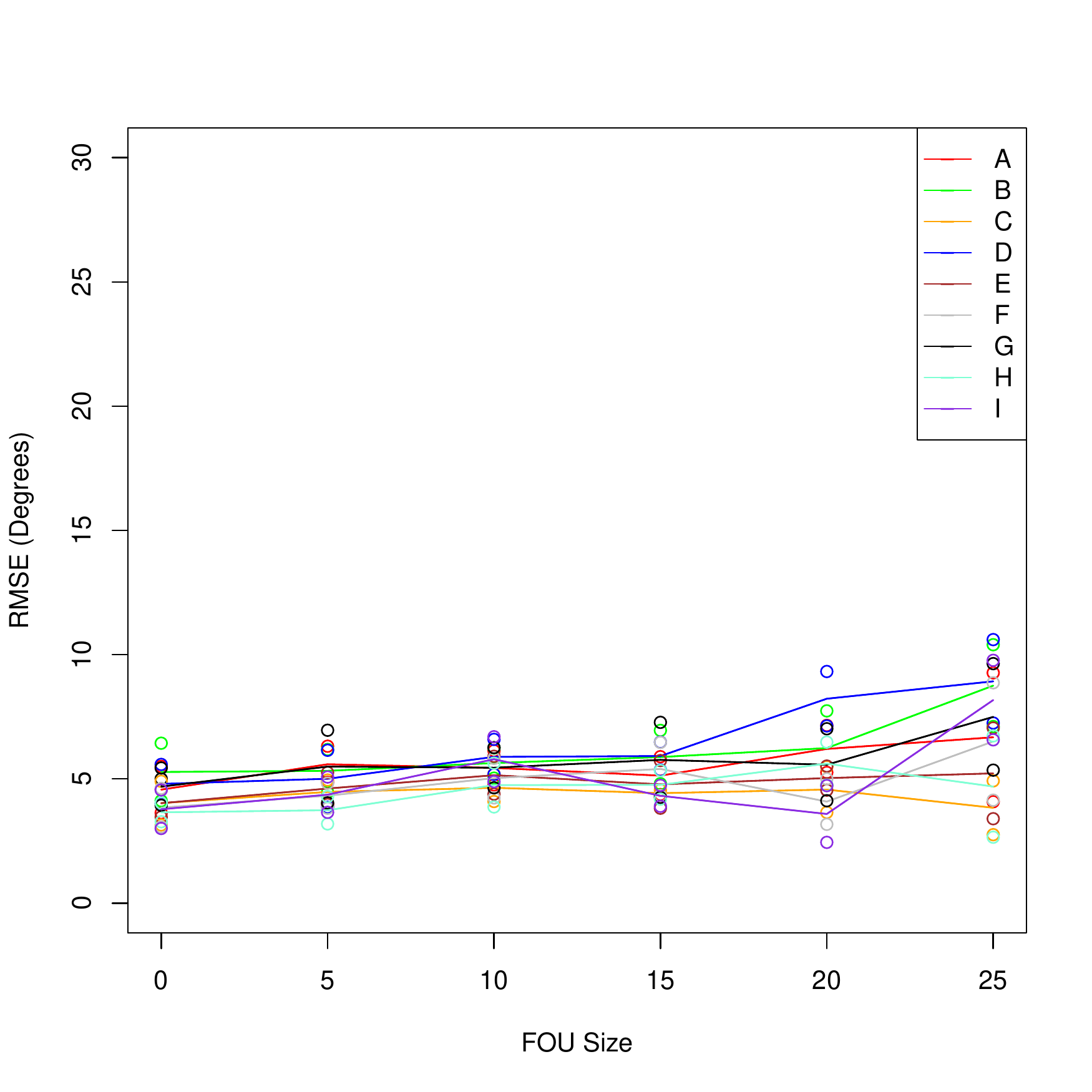}}
\subfigure[50m Vertical Movement ]{\label{fig:3_50_Config}\includegraphics[scale=0.3]{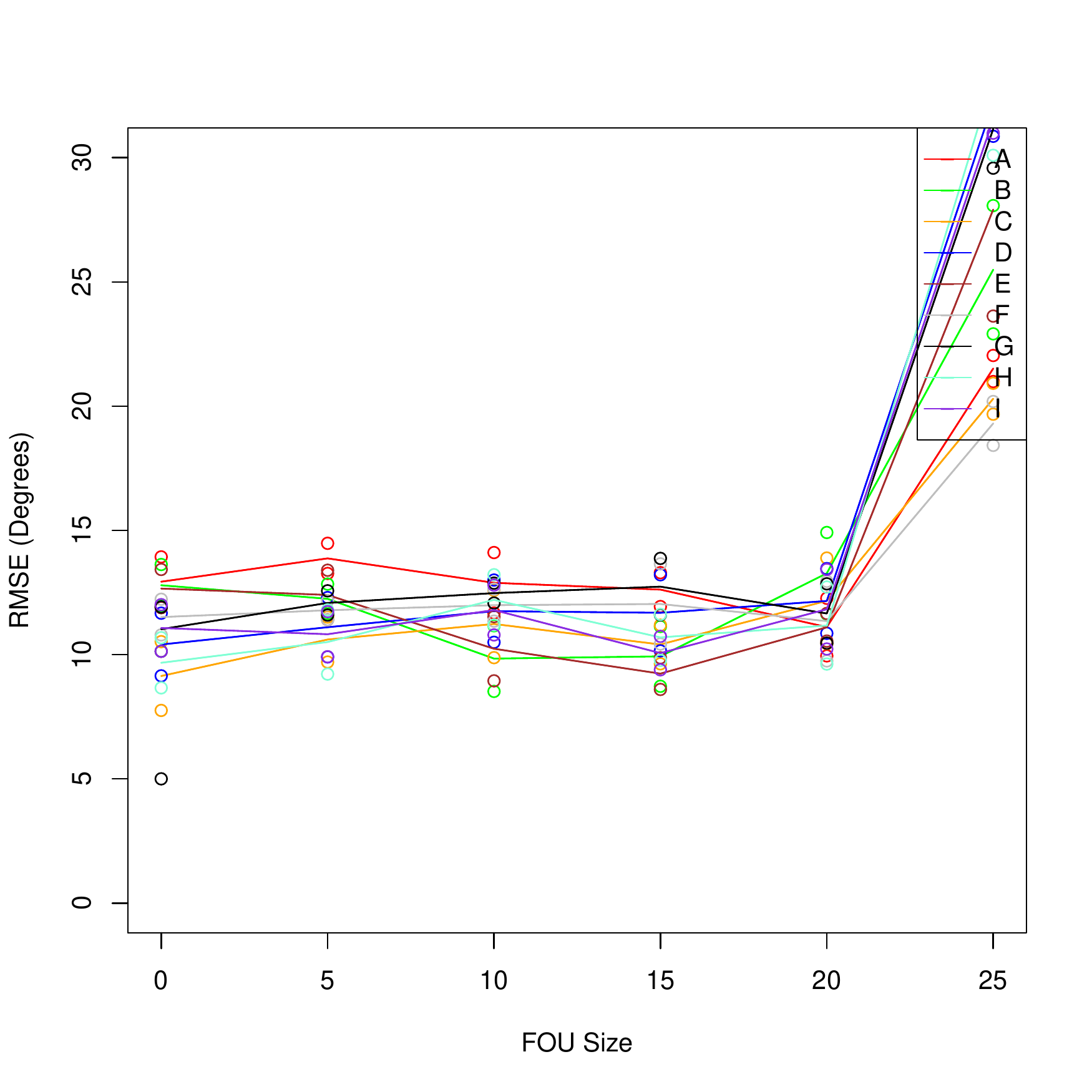}}
\subfigure[100m Vertical Movement]{\label{fig:3_100_Config}\includegraphics[scale=0.3]{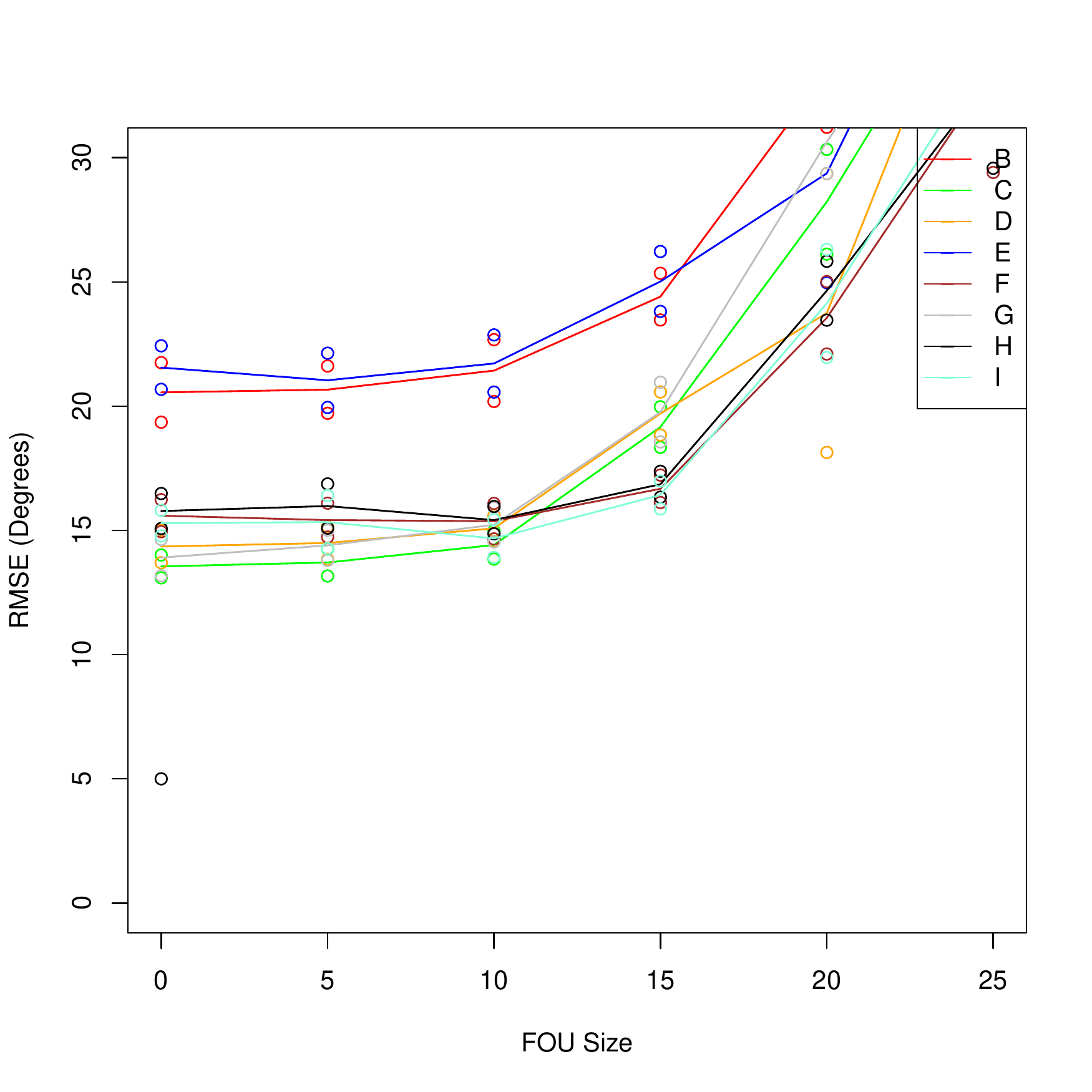}}

\caption{Double Turn Experiments RMSE Changes as FOU size increased}
\label{fig:3_Config}
\end{figure*}

\begin{figure*}[p]
\centering
\subfigure[25 RMSE Difference]{\label{fig:25_Gain}\includegraphics[scale = 0.3]{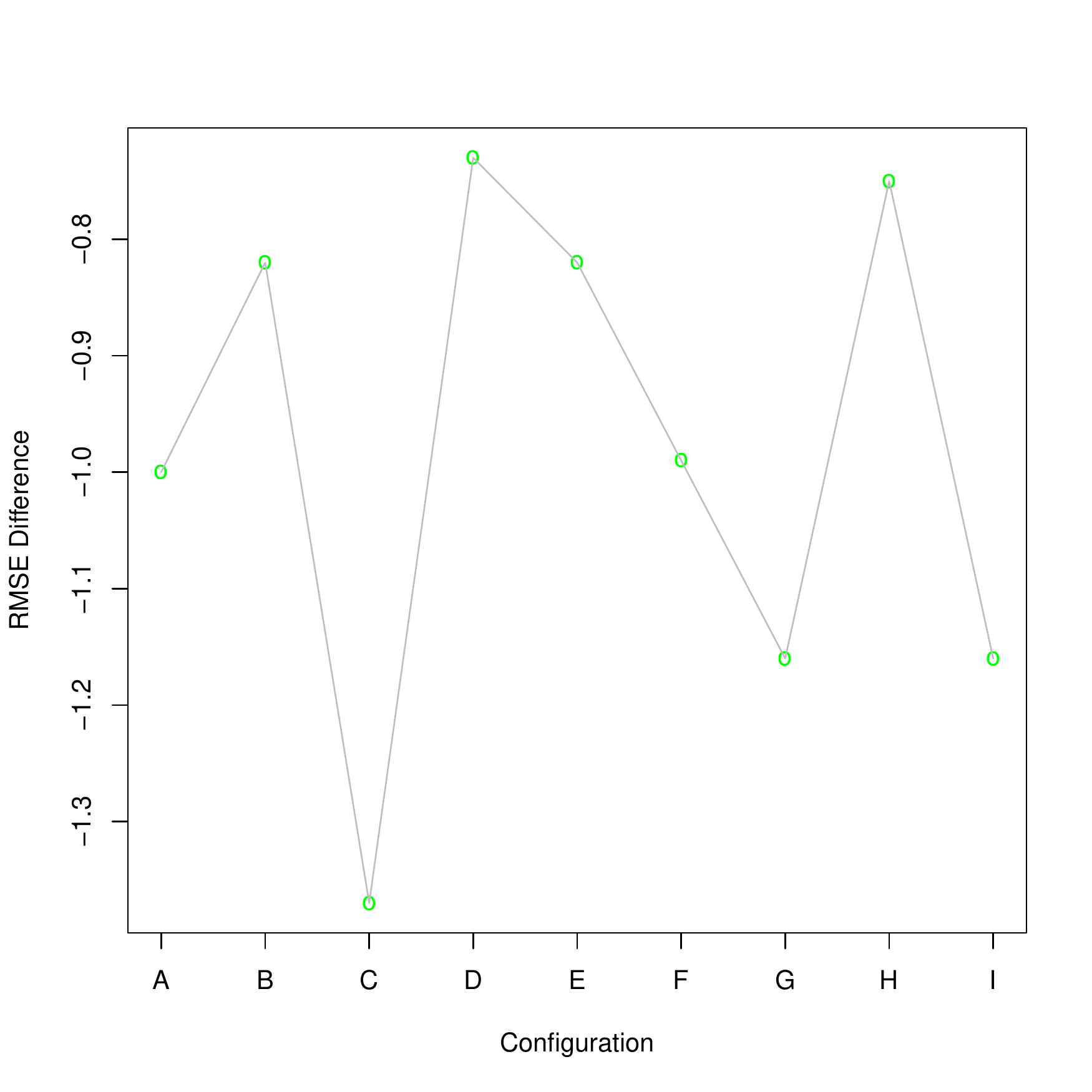}}
\subfigure[50 RMSE Difference]{\label{fig:50_Gain}\includegraphics[scale=0.3]{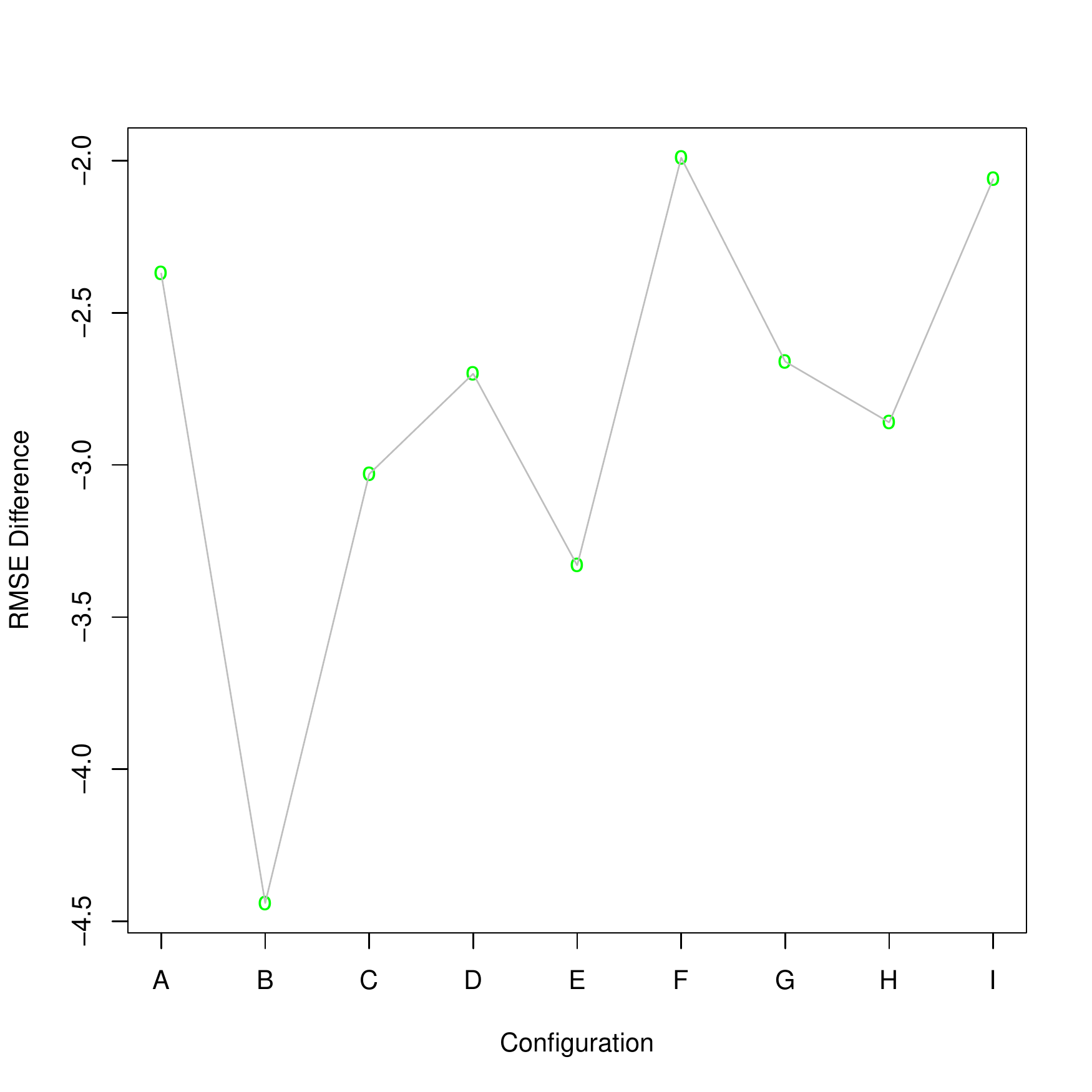}}
\subfigure[100 RMSE Difference]{\label{fig:100_Gain}\includegraphics[scale = 0.3]{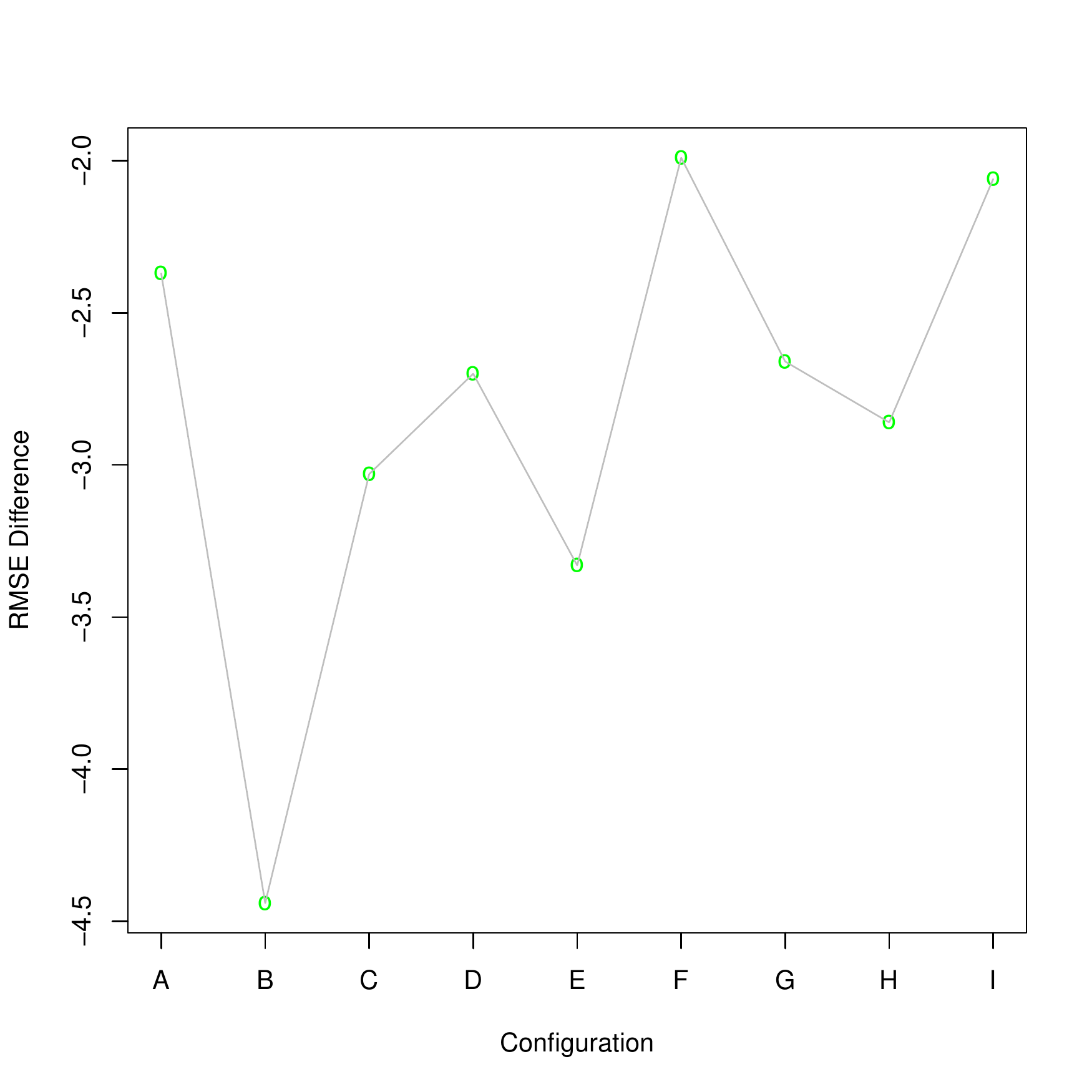}}
\caption{RMSE difference of best case type-2 controller in comparison with Type-1 controller for single turn experiments for every wind configuration.  A negative value (green circle) indicates an improvement in performance. A positive value (red cross) indicates performance decrease with a black hash indicating no change}
\label{fig:2_Gain}
\end{figure*}

\begin{figure*}[p]
\centering
\subfigure[25 RMSE Difference]{\label{fig:3_0_Gain}\includegraphics[scale = 0.3]{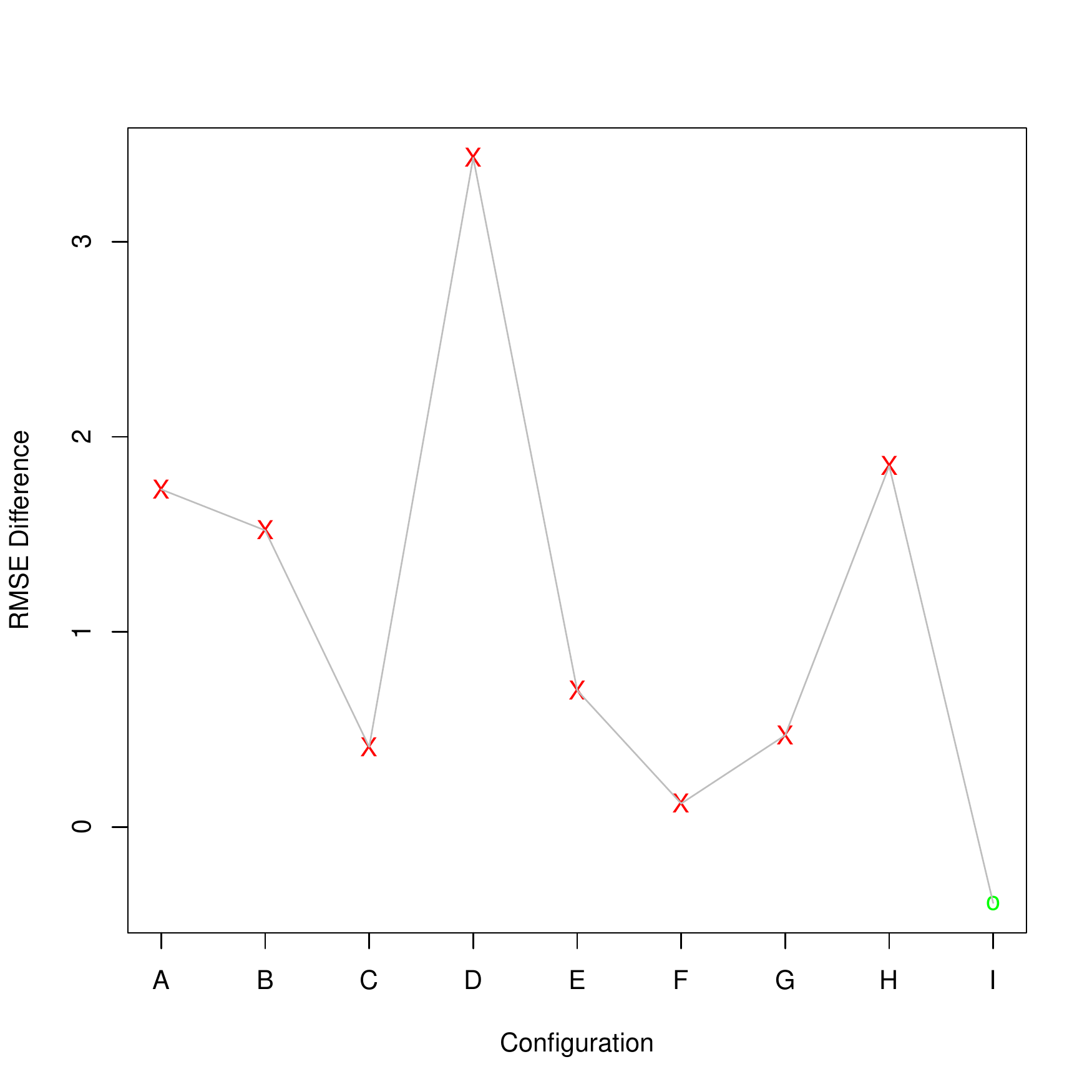}}
\subfigure[50 RMSE Difference]{\label{fig:3_50_Gain}\includegraphics[scale=0.3]{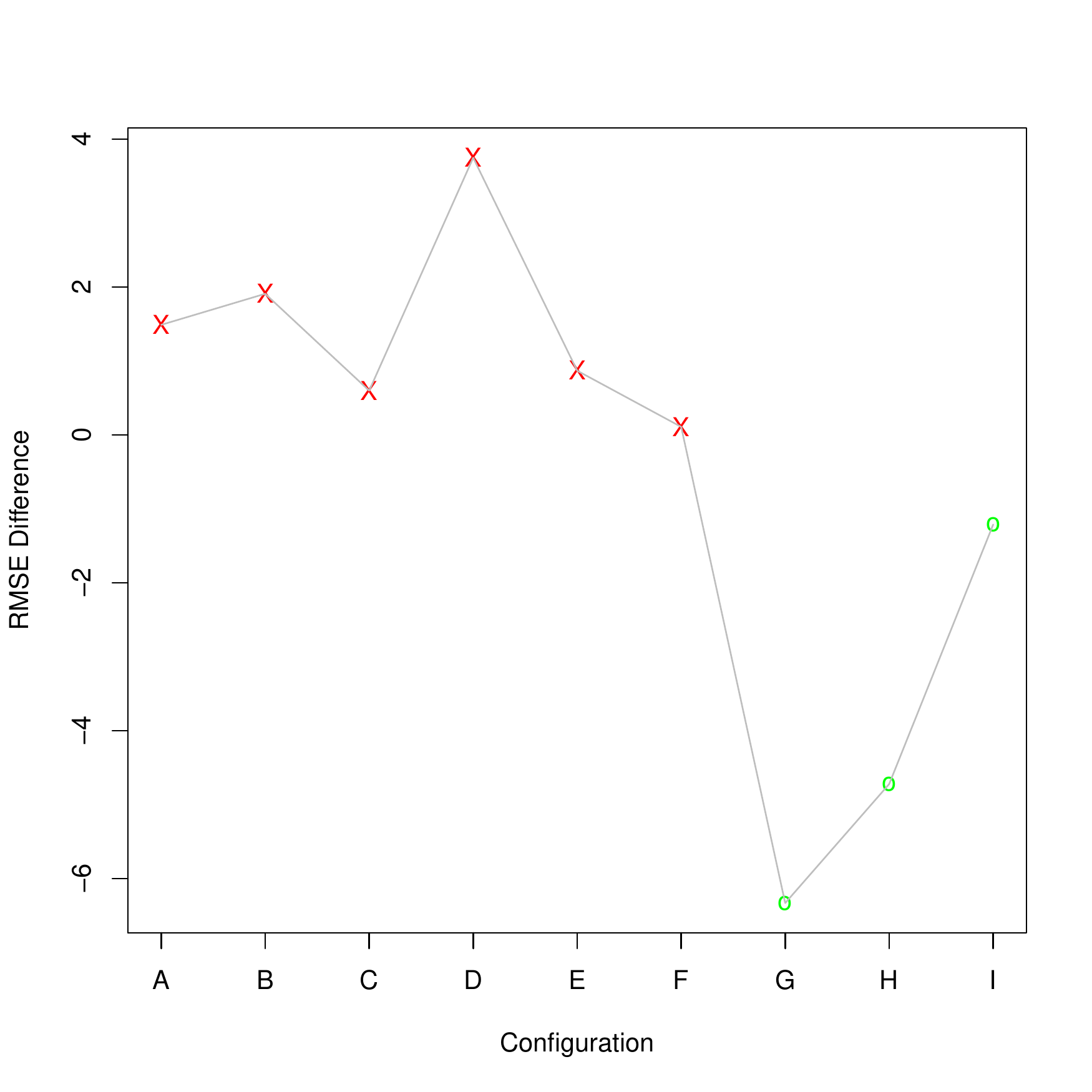}}
\subfigure[100 RMSE Difference]{\label{fig:3_100_Gain}\includegraphics[scale = 0.3]{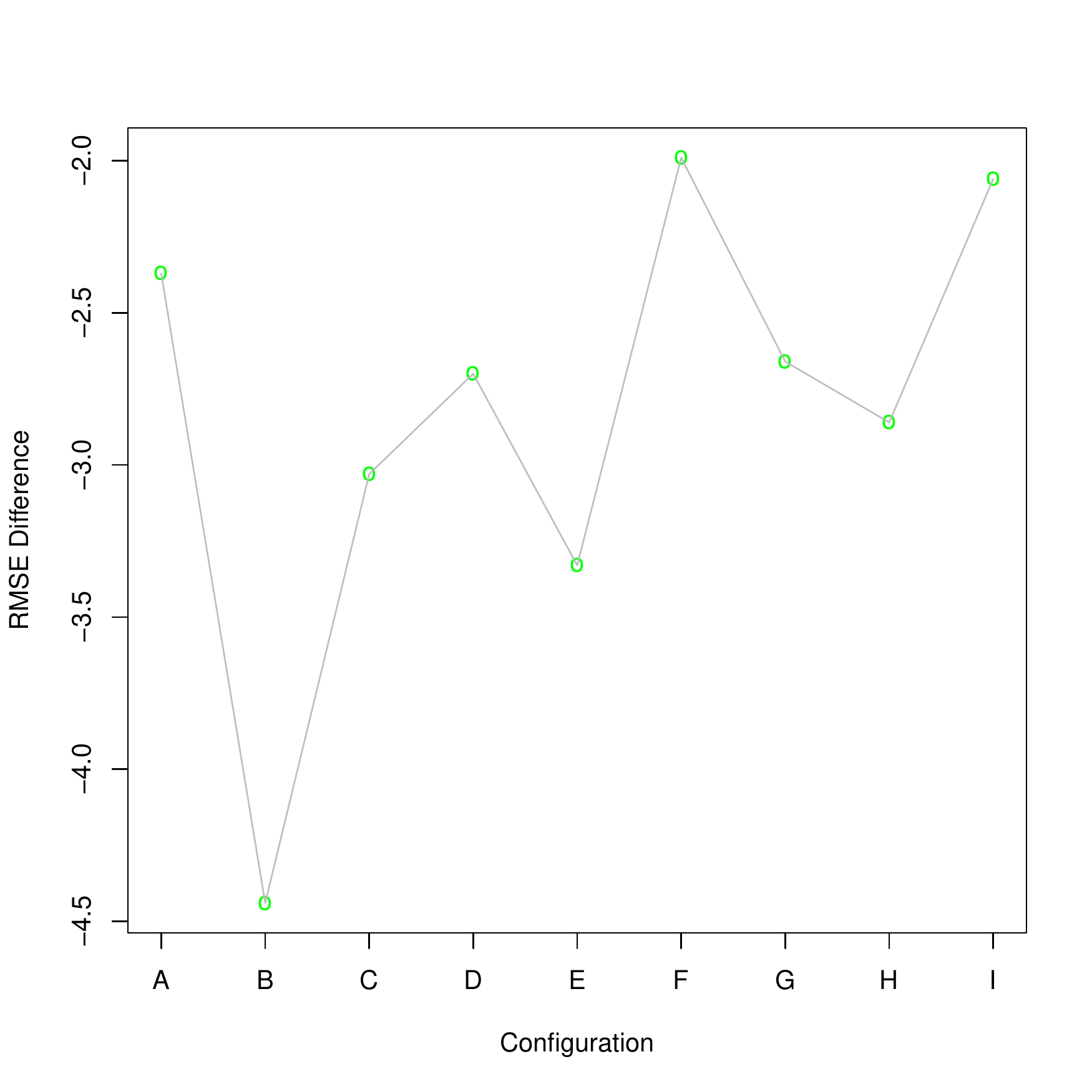}}\\

\caption{RMSE difference of best case type-2 controller in comparison with Type-1 controller for double turn experiments for every wind configuration. A negative value (green circle) indicates an improvement in performance. A positive value (red cross) indicates performance decrease with a black hash indicating no change}
\label{fig:3_Gain}
\end{figure*}

\begin{figure*}[p]
\centering
\subfigure[25m Vertical Movement Course Plot]{\label{fig:25_Plot}\includegraphics[scale = 0.3]{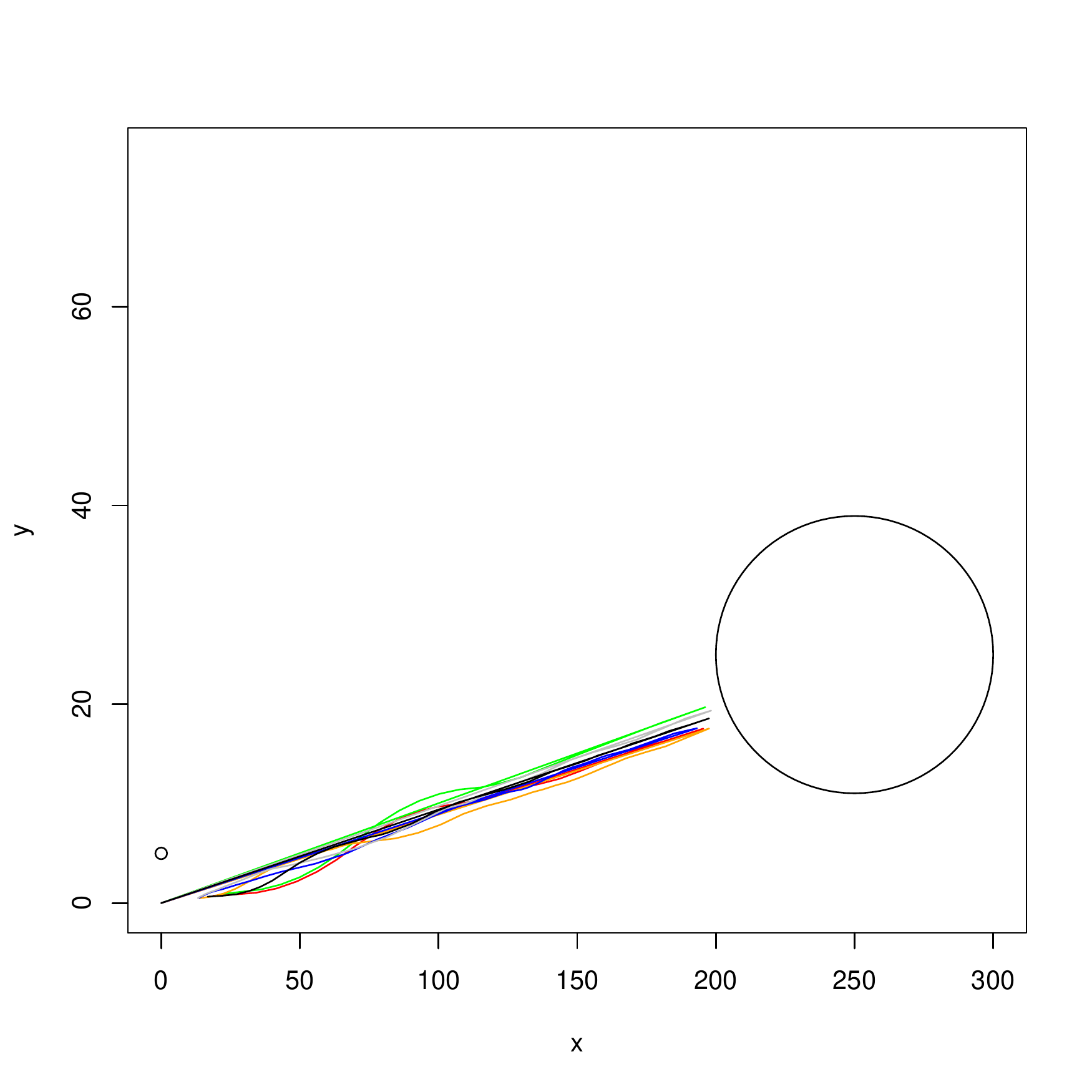}}
\subfigure[50m Vertical Movement Course Plot]{\label{fig:50_Plot}\includegraphics[scale = 0.3]{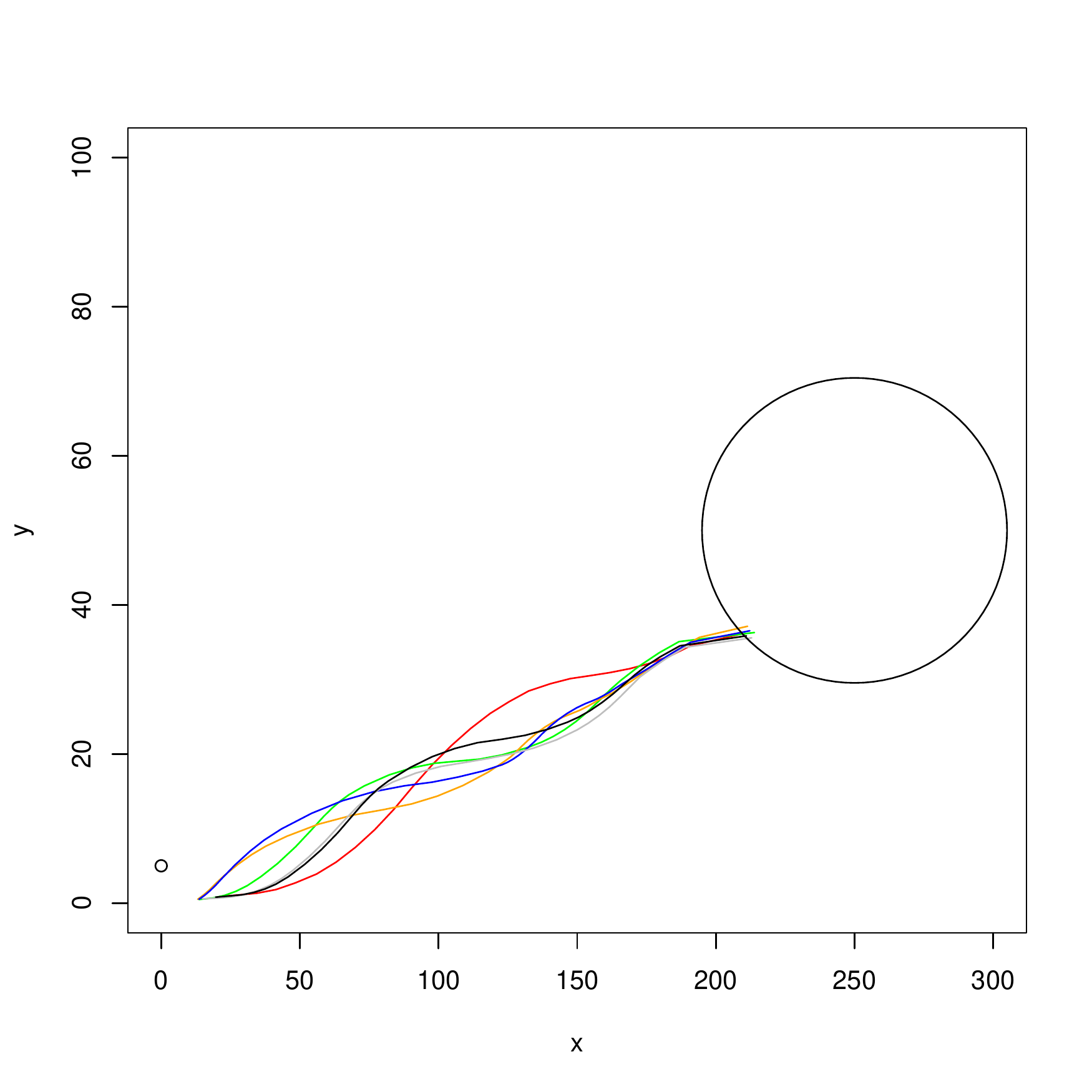}}
\subfigure[100m Vertical Movement Course Plot]{\label{fig:100_Plot}\includegraphics[scale = 0.3]{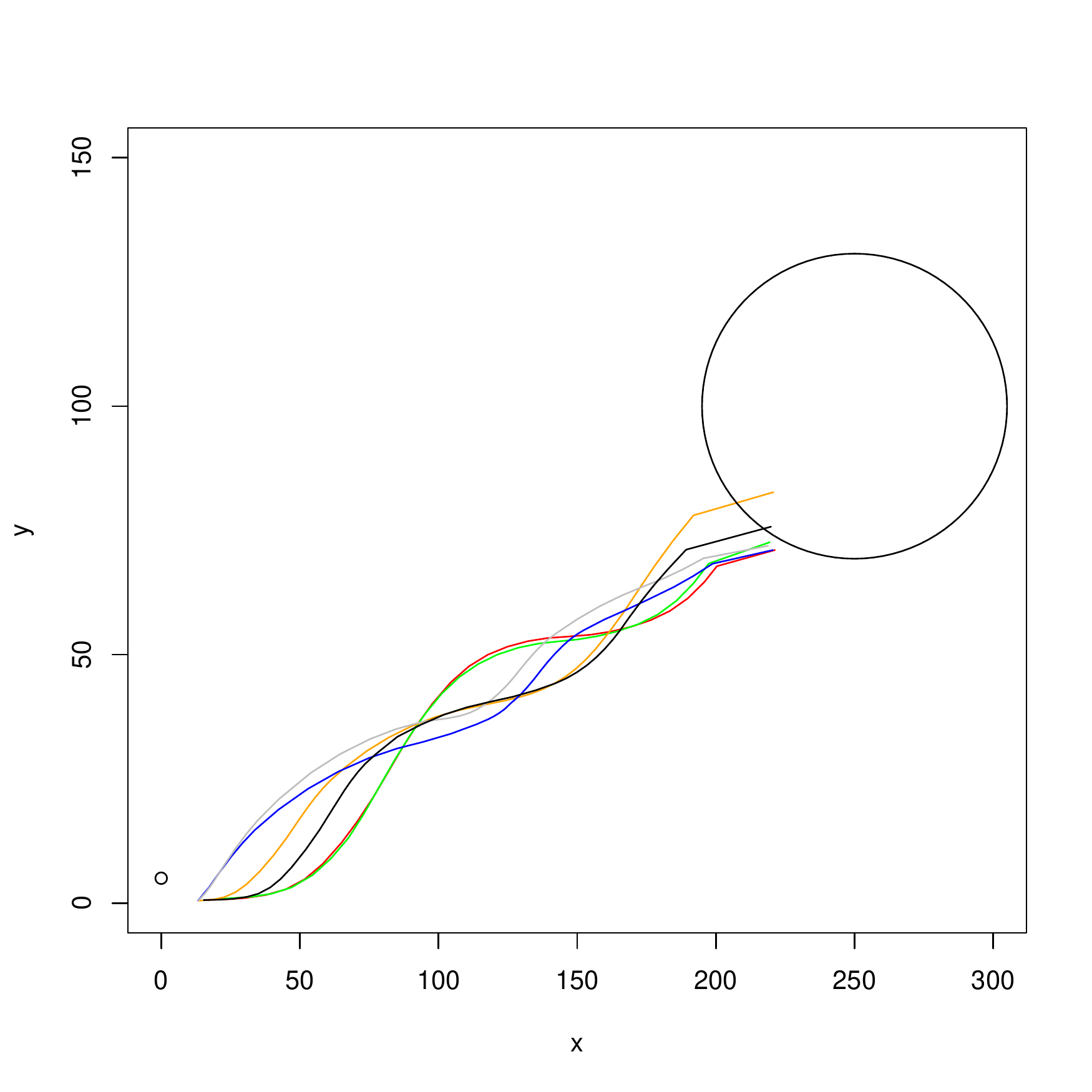}}
\caption{Example course plots for single turn experiments.  Each line represents a different wind configuration}
\label{fig:2_Course_plot}
\end{figure*}

\begin{figure*}[tb]
\centering
\subfigure[25m Vertical Movement Course Plot]{\label{fig:3_0_Plot}\includegraphics[scale = 0.3]{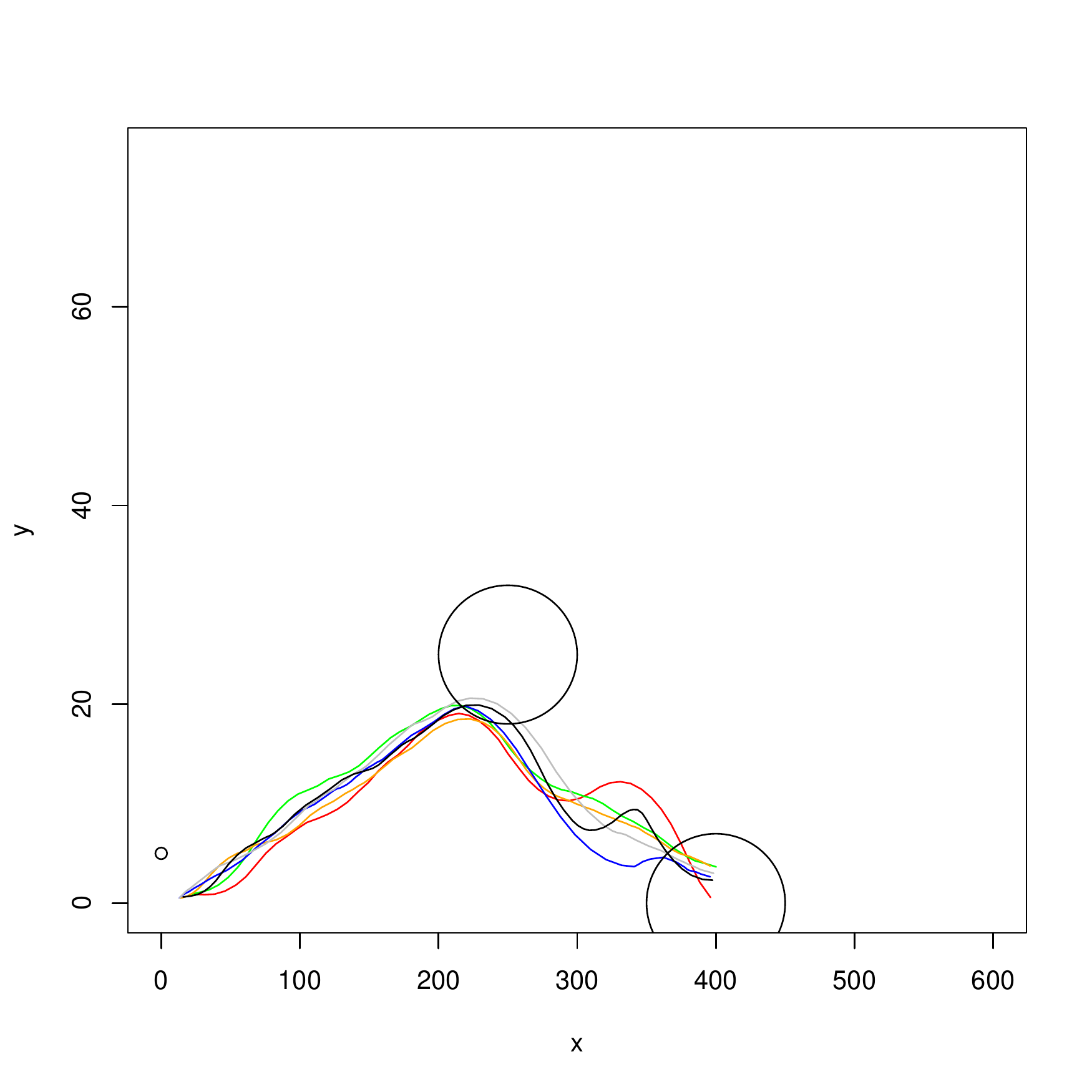}}
\subfigure[50m Vertical Movement Course Plot]{\label{fig:3_50_Plot}\includegraphics[scale = 0.3]{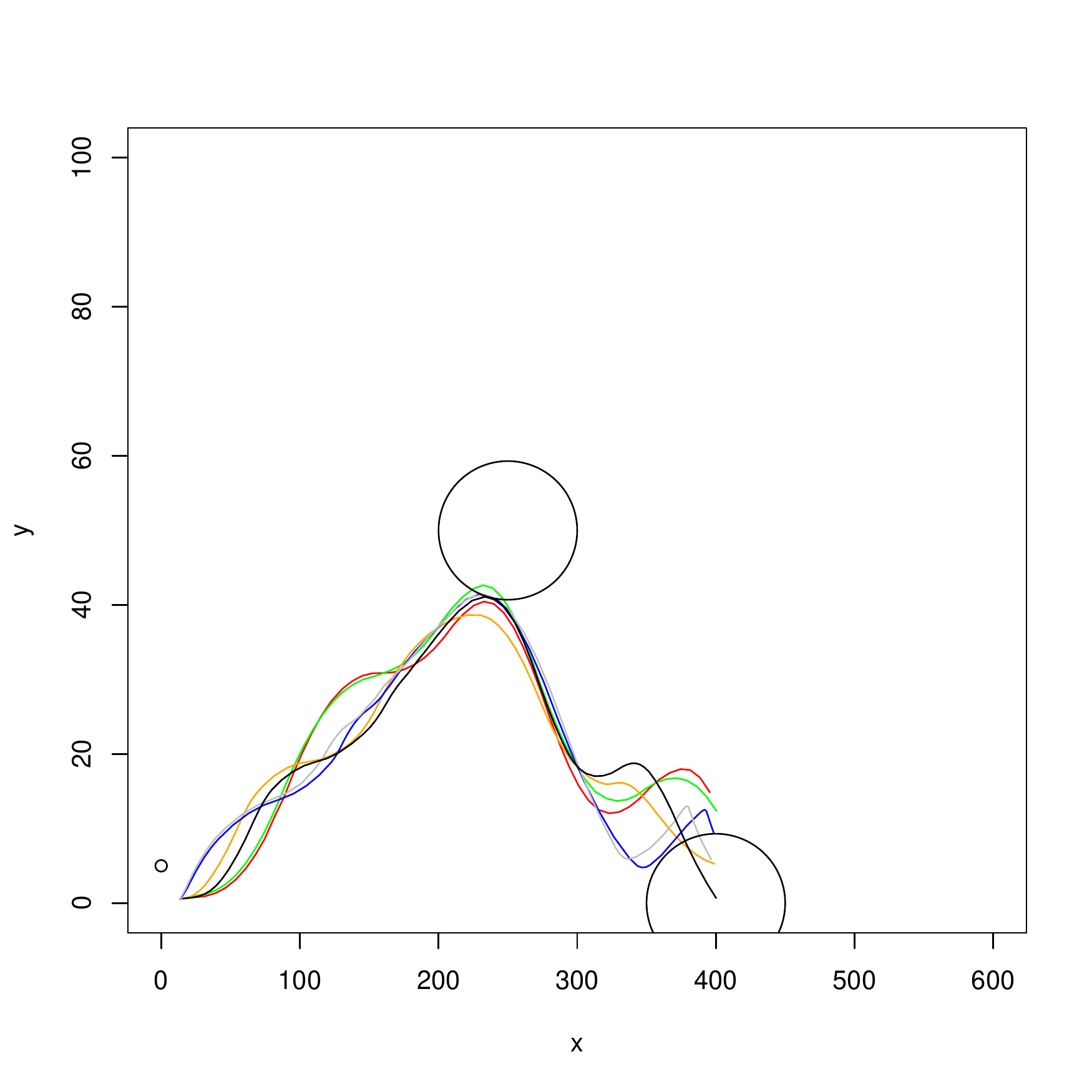}}
\subfigure[100m Vertical Movement Course Plot]{\label{fig:3_100_Plot}\includegraphics[scale = 0.3]{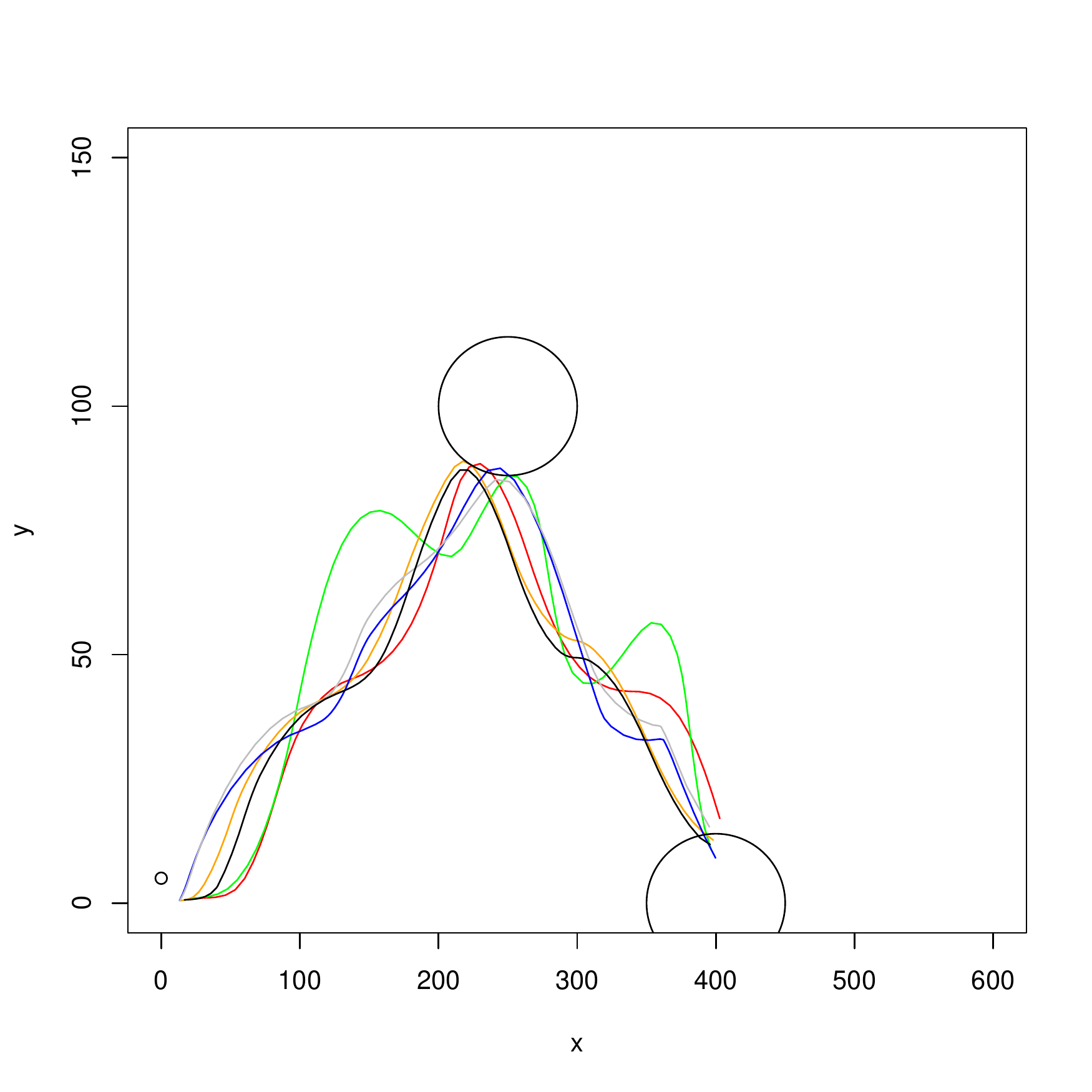}}

\caption{Example course plots for double turn experiments.  Each line represents a different wind configuration}

\label{fig:3_Course_Plot}

\end{figure*}

	We first consider Table \ref{fig:0_Point}, which shows the results of a benchmark experiment in which the majority of controllers simply maintain a straight course when the FOU size was 0 (equivalent to type-1). The average RMSE was expected at 0.0 with no statistically significant differences except with the very widest FOU sizes where performance decreases significantly as shown by the RMSE increasing in Figure \ref{fig:0_Config}. We believe that these results are caused by the fact that the controller does not need to execute any turns or course corrections in order to complete the task. This means any performance benefits/penalties a controller may have when turning do not have a chance to become apparent.
	
	The next set of data to be considered is shown in Figures \ref{fig:2_Config} and \ref{fig:3_Config}. These show how the RMSE value (on the $y$ axis) change as the FOU size is increased from 0 to 25 (as shown on the $x$ axis) for each wind configuration (each coloured line). Each course configuration is shown in a separate Figure.  In each of the figures we can see some similarities. In general the RMSE increases (signifying decreasing performance) as FOU size exceeds 20.  We can also observe that improvements in performance happen before this point, usually at a FOU size of 20 --- this is most obvious in Figure \ref{fig:50_Config}, but can also be observed in Figure \ref{fig:25_Config} and Figure \ref{fig:3_50_Config}.
	
	Figures \ref{fig:2_Gain} and \ref{fig:3_Gain} plot the difference between the type-1 controller and the best performing type-2 controller which as stated previously commonly occurs when the FOU is 20.  The $x$ axis shows the different wind configurations under test while the $y$ axis shows difference in RMSE between the type-1 and type-2 controllers.  The data for vertical movements 50m and 100m with a single turn are shown in tables \ref{tab:50_2_Diff} and \ref{tab:100_2_Point}.  It is hard to observe any obvious patterns in these plots suggesting that the noise level is not directly linked to the overall performance difference of the controller.  However in Figures \ref{fig:25_Gain} and \ref{fig:50_Gain} all points have a negative difference, shown by green circles, representing an improvement in performance over the type-1 RMSE value. In all other cases very few points show improvement over the type-1 value, shown by mostly red and black symbols.
		
	Figures \ref{fig:2_Course_plot} and \ref{fig:3_Course_Plot} show example course plots of single and double turn courses with all the various wind configurations under test represented by coloured lines and the white circles indicating way points that must be reached to complete the course.  We can see a rise in difficulty of the course with both angle and number of required turns increasing, from left to right, which in turn seems to be leading to more runs showing additional turns such as green line in Figure \ref{fig:3_100_Plot} being a good example. 
\begin{table}[bt]

\caption{RMSEs and P-Value of best performing FOU size (20 in all cases) in comparison with type-1 FOU size for single turn course configurations.}
\begin{tabular}{lllrr}
\hline\hline
Wind  & Type-1  & Type-2   & \multicolumn{1}{l}{Vertical} & \multicolumn{1}{l}{ P-Value} \\ 
Config & RMSE & RMSE & Movement & \\\hline
A & 2.72 & 1.55 & 25 & \SI{2.40e-011} {}\\ 
B & 2.82 & 1.78 & 25 & \SI{2.44e-011}{} \\ 
C & 2.60 & 1.28 & 25 & \SI{7.66e-012}{} \\ 
D & 2.81 & 1.89 & 25 & \SI{9.51e-010}{} \\ 
E & 2.58 & 1.87 & 25 & \SI{2.29e-011}{} \\ 
F & 2.16 & 1.08 & 25 & \SI{2.29e-011}{} \\ 
G & 2.17 & 1.06 & 25 & \SI{2.48e-011}{} \\ 
H & 2.67 & 1.80 & 25 & \SI{1.68e-011}{} \\ 
I & 2.24 & 0.85 & 25 & \SI{2.73e-011}{} \\ 
A & 7.00 & 3.56 & 50 & \SI{2.91e-011}{} \\ 
B & 6.76 & 3.91 & 50 & \SI{2.78e-011}{} \\ 
C & 6.51 & 3.31 & 50 & \SI{2.43e-011}{} \\ 
D & 5.99 & 3.20 & 50 & \SI{2.73e-011}{} \\ 
E & 6.98 & 4.08 & 50 & \SI{1.98e-011}{} \\ 
F & 4.86 & 2.84 & 50 & \SI{2.58e-011}{} \\ 
G & 4.85 & 2.44 & 50 & \SI{2.80e-011}{} \\ 
H & 6.49 & 3.46 & 50 & \SI{2.84e-011}{} \\ 
I & 4.82 & 2.66 & 50 & \SI{2.98e-011}{} \\ 

\hline\hline
\end{tabular}
\label{tab:2_Point_SI}

\caption{RMSEs and P-Value of best performing FOU sizes in comparison with type-1 FOU size for double turn course configurations.}

\begin{tabular}{lllrrr}
\hline\hline
Wind  & Type-1  & Type-2   & \multicolumn{1}{l}{Vertical} &FOU& \multicolumn{1}{l}{ P-Value} \\ 
Config & RMSE & RMSE & Movement & Size &\\\hline
A & 12.94 & 11.11 & 50 & 20 & \SI{1.11e-006}{} \\ 
B & 12.79 & 9.84 & 50 & 10 & \SI{4.84e-013}{} \\ 
E & 12.66 & 9.23 & 50 & 15 & \SI{3.02e-011}{} \\ 
I & 11.07 & 10.06 & 50 & 15 & \SI{1.64e-005}{} \\ 
\hline\hline
\label{tab:3_Point_SI}
\end{tabular}

\end{table}	

	Tables \ref{tab:2_Point_SI} and \ref{tab:3_Point_SI} show the P-Value obtained when comparing the type-1 controller with the best performing FOU size for each wind configuration and vertical movement combination for both single and double turn experiments respectively.  If there is no FOU size in which better performance is observed then this configuration is omitted.  We can observe firstly that there are no points in which the vertical movement is 100.  Secondly that double turn experiments have considerably fewer points than the single turn. An explanation for this will be discussed in the next section.

\section{Discussion}
\label{sec:Discussion}

	It can be observed from the results obtained and outlined in the previous section that type-2 based controllers can and do out-perform type-1 controllers in several circumstances. However this does not occur in the majority of cases. It is, in fact, more common for the performance to be similar to the type-1 value (statistically so in many but not all of cases).  
	
	If we enumerate the number of cases we find only 23 of the total of 324 (comprised of nine wind configurations, six different FOU sizes, three different vertical movement values and two different turn counts) show statistical improvement equating to approximately 8\%.  This relativity low percentage shows that those researchers who moved from type-1 to type-2 expecting a large increase in performance are more than likely to see at best the same performance but in most cases significantly worse unless considerable design effort is undertaken.
	
	Our results are supported by other works in which type-2 performance is compared with type-1 such as the work by Musikasuwan et al in \cite{GaribaldiOzen0401} where a type-1 controller outperforms, albeit by a small margin, a type-2 based controller.  While this work was more focussed on number of model parameters in each controller the essential result --- that type-1 can out perform type-2 under the correct circumstances agrees with the finding of this paper.

\begin {figure}[tb]
		\includegraphics [scale=0.5]{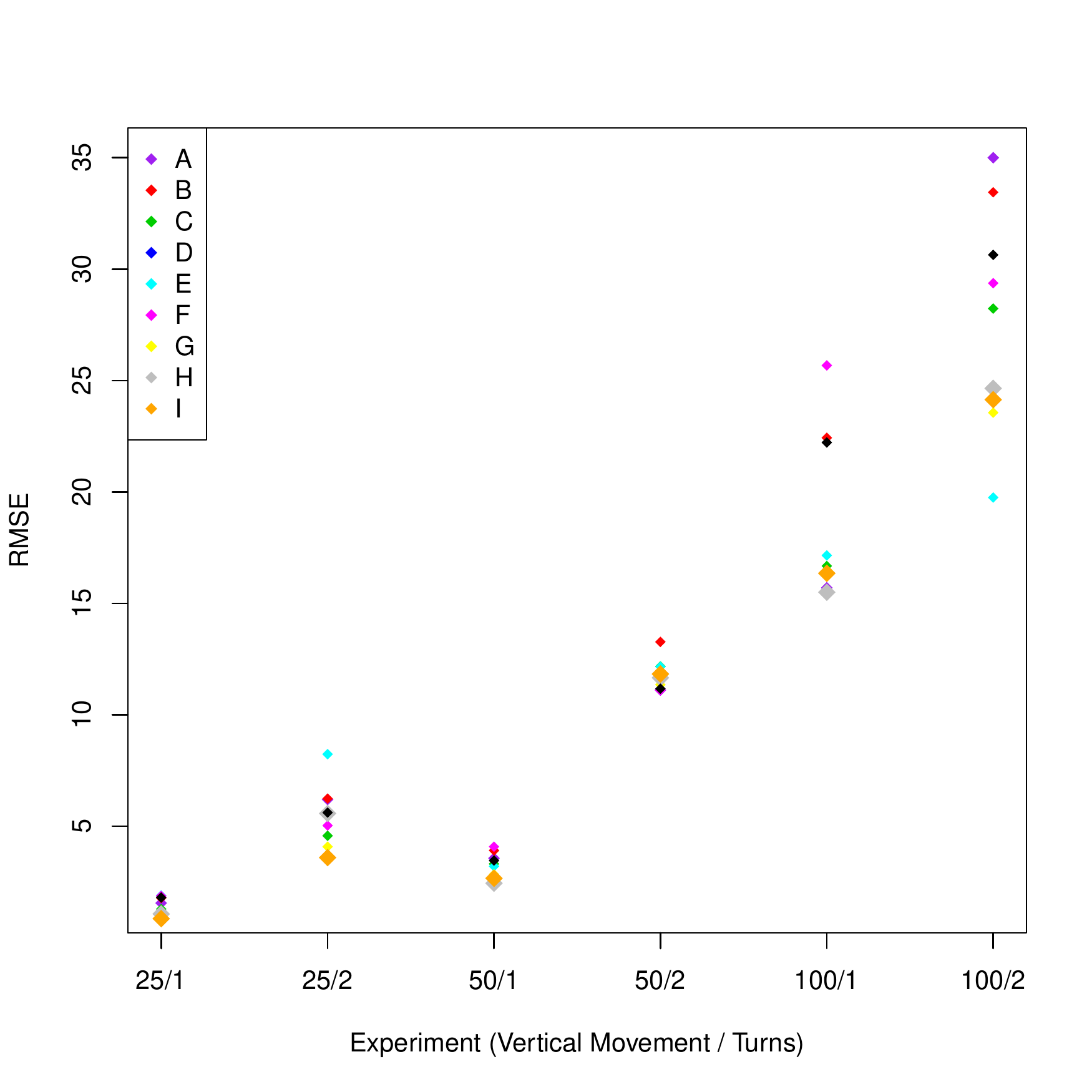}
		\caption{RMSE Values for Each wind configuration for each experiment}
 		\label{fig:ConfigNoise}

\end {figure}

	The ordering of the individual wind configurations in each of Figures \ref{fig:2_Config} and \ref{fig:3_Config} does not match with our expected hypotheses, in that the higher noise levels do not produce significantly higher RMSE values. This can be better seen in Figure \ref{fig:ConfigNoise} in which the RMSE for each wind configuration, vertical movement and turn count combination is plotted with the FOU size being held at 20.  In the majority of cases wind configuration 'B'  (red crosses) tends to have one of the the highest RMSE over the entire range of FOU sizes.  This contrasts with wind configuration 'I' (orange points) which seem to often appear at the bottom of the graph indicating the best performance.  This seems counter to what might be expected, which would be for wind configuration A to have the lowest RMSE and configuration I to have the highest (as common sense would seem to indicate that more noisy environments are more difficult to sail in).  Whether this conclusion is a general result or an artefact of the nature of this specific control problem is not yet known but will be the subject of future research.

	We also observe the spread of the results for different wind configurations increases with the course difficulty. When the vertical movement is 25 with a single turn, the results are much closer together with a difference between highest and lowers RMSE value of 1.04. This contrasts significantly with the 100-double turn experiment in which the difference is 9.98.  These results can be observed when comparing Figures \ref{fig:2_25_Plot} and \ref{fig:3_100_Plot}.  This is an expected result as with each increase in course difficulty the number of course corrections that must be done by each controller increases, meaning there is greater scope for a controller to demonstrate its improved performance (or lack thereof).
	
	The Figures \ref{fig:2_Gain} and \ref{fig:3_Gain} show that there is no obvious correlation between wind configuration (and therefore environmental noise) and the performance change achieved when moving to a type-2 controller.  This could be down to the ordering of the configurations, as defined in Table \ref{tab:Configs}.  Multiple configurations have been given an equal uncertainty score based on the assumed equal weighting of the two noise sources and this may be a faulty assumption. This also contrasts with the findings made by Sepulveda et al \cite{SEPULVEDA2007} in which type-1 and type-2 controllers are tested and the type-2 out performing the type-1 in all cases.  This occurs both with and without uncertainty and the difference in performance seems to have an increasing correlation. This suggests either the difference is down to application. Alternativly they have simply not tried as many noise configurations as we have done here meaning the differences found here have not been able to present themselves.
		
    The addition of turns to increase the difficulty of the course has a significant effect on the performance of the controllers.  It can be observed between Figures \ref{fig:2_Config} and \ref{fig:3_Config} that every value is higher in the double turn situation in comparison with the single turn.  
    
\section{Conclusions \& Future Work}
\label{sec:Conclusions_Future}

	We have shown that type-2 based controllers can and do out-perform type-1 controllers.  However, care must be taken in the design of the type-2 system, especially with regards to the size of the FOU.  Too small an FOU and the Type-2 FLC will not improve over the type-1.  Too large and it will perform worse.  In our selected application an FOU size of 20 seems to be the optimal value over a range of experimental conditions. Further work will be required to determine the reason for this value.
	
	Overall, this work shows the association between performance change and environmental noise to be considerably more complex than previously assumed.  The statement that increasing environmental noise will lead to the type-2 improving in performance compared to type-1 is not supported by the results in this paper.
	
\subsection{Future Work}

	The next step in this work will be to perform these experiments in a real world environment and to observe to what degree the findings presented here in simulation apply to a real world control problem.  Based on the results found in these experiments, generalised type-2 may be an avenue for future work.
	
\section{Acknowledgements}

	The authors would like to thank the School of Computer Science, University of Nottingham for their support in funding this paper.


\bibliographystyle {plain}
\bibliography{Journal}

\begin{thebibliography}{10}

\bibitem{Benatar2011}
N.~Benatar, U.~Aickelin, and J.~M. Garibaldi.
\newblock {A Comparison of Non-stationary , Type-2 and Dual Surface Fuzzy
  Control}.
\newblock In {\em Proceedings of IEEE International Fuzzy Systems Conference
  FUZZ-IEEE 2007}, 2011.

\bibitem{hagras2004}
H.~Hagras.
\newblock A hierachical type 2 fuzzy logic control architechture for autonomous
  mobile robots.
\newblock {\em IEEE Transactions on Fuzzy Systems}, 12, 2004.

\bibitem{NileshNKarnik1999}
N.~N. Karnik, J.~M. Mendel, and Q.~Liang.
\newblock Type 2 fuzzy logic systems.
\newblock {\em IEEE Transactions on Fuzzy Systems}, 7, 1999.

\bibitem{Mendel2006}
J.~M. Mendel, R.~I. John, and F.~Liu.
\newblock Interval type-2 fuzzy logic systems made simple.
\newblock {\em IEEE Transactions on Fuzzy Systems}, 14, 2006.

\bibitem{GaribaldiOzen0401}
S.~Musikasuwan, T.~Ozen, and J.~M. Garibaldi.
\newblock {An Investigation into the Effect of Number of Model Parameters on
  Performance in Type-1 and Type-2 Fuzzy Logic Systems}.
\newblock In {\em Proc. 10th Information Processing and Management of
  Uncertainty in Knowledge Based Systems (IPMU 2004)}, pages 1593--1600,
  Perugia, Italy, 2004.

\bibitem{Sauze2005}
C.~Sauz{\'e}.
\newblock Control software for a sailing robot.
\newblock Master's thesis, University of Wales, Aberystwyth, 2005.

\bibitem{Sauze200811}
C.~Sauz{\'e} and M.~Neal.
\newblock Design considerations for sailing robots performing long term
  autonomous oceanography.
\newblock In {\em Proceedings of The International Robotic Sailing Conference},
  2008.

\bibitem{Sauze2010}
C.~Sauz{\'e} and M.~Neal.
\newblock A neuro-endocrine inspired approach to long term energy autonomy in
  sailing robots.
\newblock In {\em Proceedings of TAROS (Towards Autonomous Robotic Systems)},
  pages 255--262, Bannf, Canada, 17 - 19 July 2010.

\bibitem{SEPULVEDA2007}
R.~Sepulveda, O.~Castillo, P.~Melin, A.~Rodriguezdiaz, and O.~Montiel.
\newblock {Experimental study of intelligent controllers under uncertainty
  using type-1 and type-2 fuzzy logic}.
\newblock {\em Information Sciences}, 177(10):2023--2048, May 2007.

\bibitem{Stelzer-20081}
R.~Stelzer.
\newblock Robotic sailing: Overview.
\newblock {\em OGAI Journal (Oesterreichische Gesellschaft fuer Artificial
  Intelligence)}, 27(2):2--3, June 2008.

\bibitem{Stelzer-2008}
R.~Stelzer and T.~Pr{\"o}ll.
\newblock Autonomous sailboat navigation for short course racing.
\newblock {\em Robotics and Autonomous Systems}, 56(7):604--614, July 2008.

\bibitem{Wagner2011}
C.~Wagner and H.~Hagras.
\newblock Employing zslices based general type-2 fuzzy sets to model multi
  level agreement.
\newblock In {\em Advances in Type-2 Fuzzy Logic Systems (T2FUZZ), 2011 IEEE
  Symposium on}, pages 50 --57, april 2011.

\bibitem{Wu2011a}
D.~Wu.
\newblock {An Interval Type-2 Fuzzy Logic System Cannot Be Implemented by
  Traditional Type-1 Fuzzy Logic Systems}.
\newblock In {\em World Conference on Soft Computing}, number~x, 2011.

\bibitem{Wu2006}
D.~Wu and W.~W. Tan.
\newblock {A simplified type-2 fuzzy logic controller for real-time control.}
\newblock {\em ISA transactions}, 45(4):503--16, October 2006.

\bibitem{Zadeh1975}
L.~A. Zadeh.
\newblock The concept of a linguistic variable and its application to
  approximate reasoning i.
\newblock {\em Information Sciences}, pages 199--249, 1975.

\end{thebibliography}



\end{document}